\documentclass[preprint,authoryear,a4paper]{elsarticle}
\usepackage{geometry}
\usepackage{multicol}

\usepackage{color,soul}

\usepackage{url}
\usepackage[hidelinks]{hyperref}
\hypersetup{
    colorlinks   = true,
    urlcolor     = blue,
    linkcolor    = blue,
    citecolor    = blue
}

\emergencystretch=\maxdimen
\usepackage{graphicx}
\graphicspath{{./pics/}}
\DeclareGraphicsExtensions{.pdf,.png,.jpg}
\usepackage{placeins}
\usepackage{subcaption}
\usepackage{float}
\usepackage{caption}

\usepackage{booktabs}
\usepackage[table,xcdraw,dvipsnames]{xcolor}
\usepackage{color,xcolor}
\usepackage{graphicx}
\usepackage{adjustbox}
\usepackage{multirow}
\usepackage{tabularx}
\usepackage{utfsym}

\usepackage{amssymb}
\usepackage{gensymb}
\usepackage[mathscr]{euscript}
\usepackage{bm}
\usepackage{pifont}

\usepackage{mathtools}
\usepackage{nccmath}
\usepackage{amsmath}
\usepackage{braket}
\usepackage[ruled]{algorithm2e}
\usepackage{algpseudocode}

\usepackage{lineno,hyperref}

\usepackage{lipsum}

\journal{}

\begin{document}

\begin{frontmatter}
	\title{BCE-Net: Reliable Building Footprints Change Extraction based on Historical Map and Up-to-Date Images using Contrastive Learning}

    \author[addr1]{Cheng Liao}
    \author[addr1]{Han Hu\corref{cor}}
    \author[addr1]{Xuekun Yuan}
    \author[addr2]{Haifeng Li}
    \author[addr3]{Chao Liu}
    \author[addr3]{Chunyang Liu}
    \author[addr4]{Gui Fu}
    \author[addr1]{Yulin Ding}
    \author[addr1]{and Qing Zhu}
    \cortext[cor]{Corresponding Author: han.hu@swjtu.edu.cn}
    
    \address[addr1]{Faculty of Geosciences and Environmental Engineering, Southwest Jiaotong University, Chengdu, China}
    \address[addr2]{School of Geosciences and Info-Physics, Central South University, Changsha, China}
    \address[addr3]{School of Spatial Information and Geomatics Engineering, Anhui University of Science and Technology, Huainan, China}
    \address[addr4]{School of Construction Equipment, Guizhou Polytechnic of Construction, Guiyang, China}

    \begin{abstract}
        Automatic and periodic recompiling of building databases with up-to-date high-resolution images has become a critical requirement for rapidly developing urban environments. However, the architecture of most existing approaches for change extraction attempts to learn features related to changes but ignores objectives related to buildings. This inevitably leads to the generation of significant pseudo-changes, due to factors such as seasonal changes in images and the inclination of building façades. To alleviate the above-mentioned problems, we developed a contrastive learning approach by validating historical building footprints against single up-to-date remotely sensed images. This contrastive learning strategy allowed us to inject the semantics of buildings into a pipeline for the detection of changes, which is achieved by increasing the distinguishability of features of buildings from those of non-buildings. In addition, to reduce the effects of inconsistencies between historical building polygons and buildings in up-to-date images, we employed a deformable convolutional neural network to learn offsets intuitively. In summary, we formulated a multi-branch building extraction method that identifies newly constructed and removed buildings, respectively. To validate our method, we conducted comparative experiments using the public Wuhan University building change detection dataset and a more practical dataset named SI-BU that we established. Our method achieved F1 scores of 93.99\% and 70.74\% on the above datasets, respectively. Moreover, when the data of the public dataset were divided in the same manner as in previous related studies, our method achieved an F1 score of 94.63\%, which surpasses that of the state-of-the-art method. We will open-source the code and model to reproduce the results for further research at \url{ https://github.com/liaochengcsu/BCE-Net}.
        
    \end{abstract}
    \begin{keyword}
    	Building update \sep Change detection \sep Semantic segmentation \sep Contrastive learning
    \end{keyword}
\end{frontmatter}


\section{Introduction}
\label{s:introduction}
Efficient and accurate acquisition of up-to-date building footprints is required for various applications, such as land-use investigation and illegal-building monitoring \citep{guo2017mining,lv2021land,vieira2012object}. 
Existing building databases collated by official cadastral departments or crowd-sourcing via Open Street Map \citep{fan2014quality,franklin2011crowddb} have been established gradually using multi-temporal and multi-source data.
Thus, there is an urgent need for these building databases to be updated using uniform data sources, especially databases that cover rapidly developing urban environments \citep{lands_department_3d_2022,accucities_3d_2022}. 
High-resolution satellite images appear the most suitable type of uniform data sources to fulfill the above objective at regular time intervals, especially as they are both time- and cost-efficient.

Extraction of building changes is the first step towards complete updating of historical building polygons, and the identification of building changes in remotely sensed images has been widely researched over the past decades \citep{lv2021land,afaq2021analysis,sefrin2020deep}. 
A commonly employed strategy identifies changes by comparison of bi-temporal co-registered images. The Siamese network with shared encoder parameters is often used to learn discriminable features and extract pixel-wise changes \citep{lv2021land,liu2021building,shi2020change}.
Another strategy is to validate the changing state of specific image patches by examination of existing building polygons \citep{guo2017mining,zhu2008change,zhu2021depth}, which is an object-level classification task.
 Although existing convolutional neural network (CNN) methods have made impressive progress on extracting building changes via image-to-image and map-to-image approaches, such as those described above \citep{chen2021remote,zhang2022swinsunet}, the problems they encounter (as detailed below) highlight the need for a more practical and reliable method for the extraction of building changes.

\begin{figure}[H]
	\centering
	\includegraphics[width=0.6\linewidth]{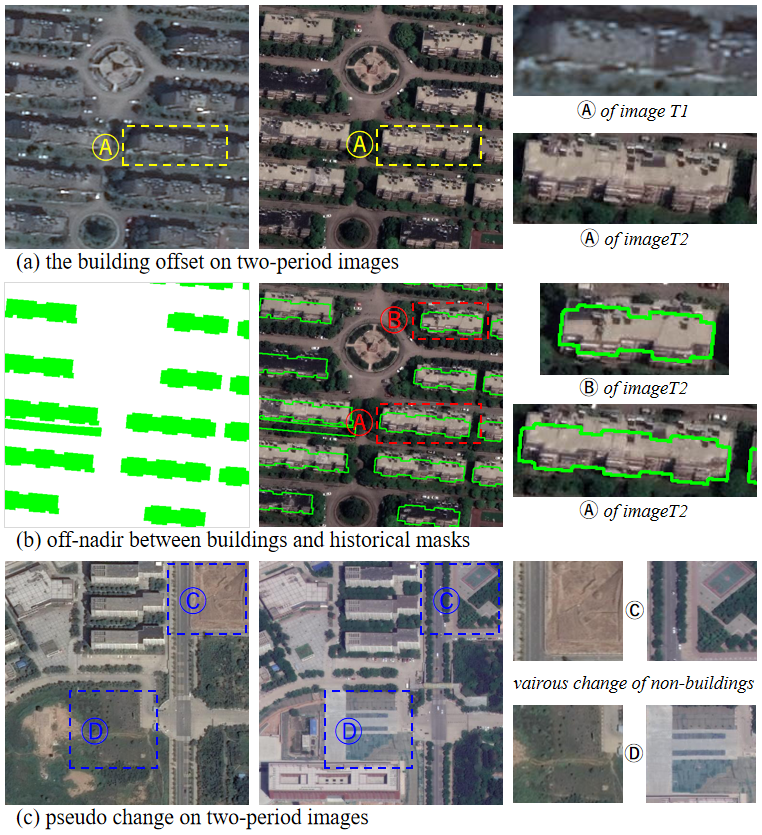}
	\caption{Examples of two existing problems that are unavoidable when using existing building change-detection approaches: off-nadir and pseudo-change problems in two-period data. The first two columns contain images and building masks in two periods, respectively. The last column shown the details of the boxed area(s) in the previous image(s).}
	\label{fig:fig1}
\end{figure}

\paragraph{1) Geometric inconsistencies of building instances}
Because of differences in the incident angles of multi-temporal satellite images, building rooftops and footprints in images are not exactly aligned.
These offsets are varying and unpredictable due to the different locations and heights of buildings, which typically results in change misidentification around unchanged buildings, as shown in Figure \ref{fig:fig1} (a).
In addition, the viewing angle from images can result in significant off-nadir \citep{zhong2015building,girard2018aligning} between building representations in images and preexisting footprints, as shown in Figure \ref{fig:fig1} (b). These unavoidable geometric inconsistencies hinder the accurate detection of building changes.

\paragraph{2) Ignorance of building semantics}
Existing image-to-image strategies learn to discriminate features based on the pixel-wise differences of bi-temporal images, which reveal general changes. 
Thus, these strategies ignore the context of pixels and neglect prior characteristics of buildings, such as their specific structures and textures.
Furthermore, the presence of different types of illumination and sensors, and various non-building changes in backgrounds \citep{chen2020spatial,sun2022fair1m}, means that these strategies generate numerous pseudo-changes, as shown in Figure \ref{fig:fig1} (a) and (c). 
To avoid the impact of such pseudo-changes, and given that building changes only occur where a building exists at one of the two-period data, strategies are needed that focus more than current strategies on the contrastive features of buildings and non-buildings and therefore better exploit potential a priori information.

\begin{figure}[H]
	\centering
	\includegraphics[width=0.55\linewidth]{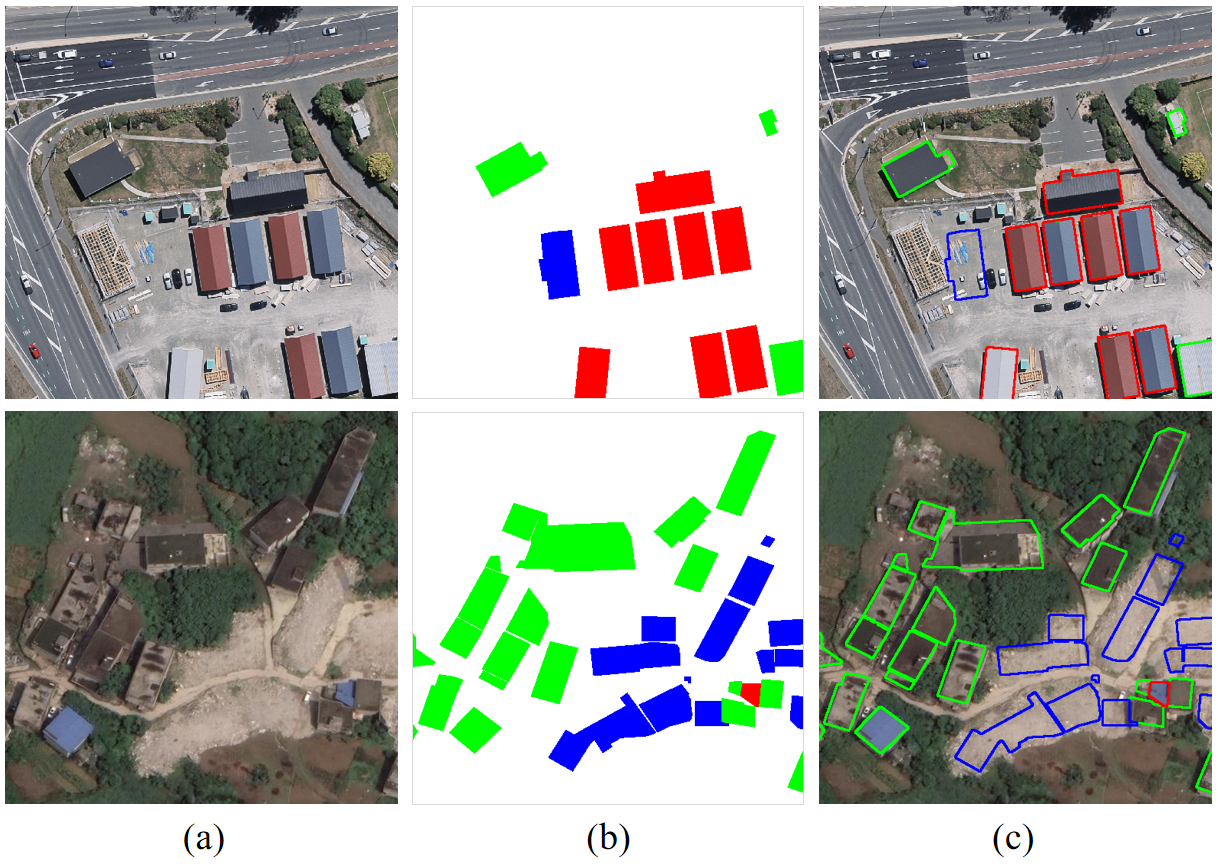}
	\caption{Samples of the WHU-CD and constructed SI-BU datasets. (a) Up-to-date images. (b) Fully labeled change state obtained by combining the images and corresponding historical building footprints. The red, blue, and green shapes represent newly constructed, removed, and unchanged buildings, respectively. (c) Overlay of images and labels.}
	\label{fig:fig2}
\end{figure}

Accordingly, we developed the building change extraction network (BCE-Net), which is a multi-task segmentation framework that combines a single up-to-date image with historical building footprints to simultaneously identify newly constructed and removed buildings.
BCE-Net overcomes geometric inconsistencies caused by off-nadir problems by employing a deformable convolutional network (DCN) to directly validate non-aligned building polygons and images.
Furthermore, BCE-Net exploits an instance-level contrastive learning strategy based on potential consistency between buildings and diversity between buildings and backgrounds, thereby achieving more robust and reliable building change extraction than current strategies.
In addition, we constructed a public building change extraction dataset named SI-BU that is more practical than current datasets and used it to evaluate the performance of BCE-Net. 
This afforded a new benchmark for comparison of related research. The SI-BU dataset is composed of single temporal satellite imagery and fully labeled ground truth based on corresponding historical building footprints.
Samples of the Wuhan University building change detection (WHU-CD) dataset and our SI-BU dataset are shown in Figure \ref{fig:fig2}.

The main contributions of this study are as follows. 
(1) We designed a novel building-change extraction framework, BCE-Net, that simultaneously extracts newly constructed and removed buildings by combining up-to-date images with historical building masks and thus has the potential to be used for updating building databases.
(2) We devised an instance-level contrastive learning strategy that exploits potential global consistency and diversity between buildings and backgrounds, thereby improving the discrimination and robustness of building change extraction.

Crucially, unlike current building change-extraction approaches based on pixel-wise differences between bi-temporal images, BCE-Net avoids the problem of pseudo-change and also significantly alleviates the offset between buildings in two periods.
The remainder of this paper is organized as follows. In section \refeq{s:relatedworks}, we briefly review previous studies in this area. In section \ref{s:methodology}, we describe the details of BCE-Net. In section \ref{s:experiments}, we detail the experiments we conducted to evaluate the performance of BCE-Net, and then analyze the results. Section \ref{s:conclusion} concludes this paper.

\section{Related works}
\label{s:relatedworks}
Advanced architectures, such as the transformer \citep{zhang2022swinsunet,he2022transformer} and graph-based networks \citep{he2022multi} are successfully applied in the remote sensing interpretation and obtain a significant improvement. However, there are still some challenges that need to be resolved for building change extraction. Here, we review previous studies on building change detection, contrast learning, and transfer learning.

\paragraph{1) Building change detection}
Building change detection aims to identify newly constructed buildings and removed buildings by analyzing differences in images of the same area over time. \citet{long2015fully} used a fully convolutional network to determine the difference between paired images. \citet{peng2019end} designed an end-to-end change-detection framework based on a semantic segmentation network that extracts multi-scaled change features from concatenated bi-temporal remotely sensed imagery. \citet{liu2020deep} replaced the normal convolution layers of U-Net \citep{ronneberger2015u} with depth-wise separable convolution layers, which afforded a lighter network that extracts changes from concatenated images. Rather than concatenating paired images, \citet{zhang2017separate} employed segmentation networks to extract buildings separately from different temporal images and established spatial correspondence for building change detection. Similarly, \citet{zhang2020deeply} designed a two-stream architecture to extract features from bi-temporal imagery and obtain differential features through a difference discrimination network that improved change detection.

\citet{zhan2017change} designed the Siamese network, which uses parameter-shared encoders to directly extract changed features from paired images and uses a contrast learning loss to optimize the distance of pixel-wise features. \citet{daudt2018fully} extended this approach by developing a fully convolutional Siamese network with skip connections, thereby enhancing change-detection performance. These studies have demonstrated that Siamese networks greatly enhance the efficiency and accuracy of change detection.

\citet{zheng2021clnet} recognized the variation in the scale of buildings by extracting multi-scale features via a cross-layer convolutional network and aggregating multi-level context, which improved the accuracy of detected building changes. Similarly, the DSA-Net \citep{ding2021dsa} introduced spatial attention and an atrous spatial pyramid-pooling module to integrate multi-scaled contextual information, thereby achieving sufficient feature extraction. \citet{huang2021multiple} devised the mutual-attention Siamese network, which embeds an attention module into the Siamese network to realize adaptive selection and fusion of multi-scale features. The SRCDNet \citep{liu2021super} enhanced multi-scale features by employing stacked attention modules and used an adversarial learning-based super-resolution module to decrease the resolution difference between bi-temporal images. 

To improve the integrity of recognized changes, \citet{peng2020optical} developed a deep dilated convolutional neural network (DDCNN) that uses attention layers in its up-sampling blocks to model the internal correlation between high-level and low-level features. This DDCNN focuses on differentiating features of image pairs and generates more accurate building change results than previous CNNs. The STPNet \citep{yang2021spatio} designed a spatiotemporal processing module to capture long-range correlations and bi-directional contextual semantics based on existing Siamese architecture to achieve precise change recognition.

The EGRCNN \citep{bai2021edge} integrated discriminative features and structural priors of building edges to refine the boundaries of extracted results for improving the accuracy of boundary detection of changed buildings. The HFA-Net \citep{zheng2022hfa} applied a spatial attention-based enhancement module to strengthen high-frequency features for boundary-refined building change extraction. \citet{hu2021cascaded} and \citet{wang2022learning} have considered the inconsistency between bi-temporal data in investigating the prediction of the instance-level offset vectors between buildings' footprints and roofs. 

However, although the off-nadir problem has recently attracted the attention of the remote sensing community, it is not feasible to automatically correct the various building offsets present in large-scale paired remotely sensed imagery. Moreover, pseudo-changes that result from aberrant detection of the environment remain the greatest problem that must be solved to realize reliable detection of building changes.

\paragraph{2) Contrast learning}

Contrast learning \citep{chen2020simple,khosla2020supervised} improves the robustness and quality of learned features by increasing the distance between positive and negative samples and decreasing the distance within pairs of positive or negative samples. In addition, contrastive learning has been widely utilized in change detection tasks to alleviate the problems of pseudo-changes and the imbalance between positive and negative samples, thereby facilitating the extraction of robust discriminative features \citep{zhan2017change}. 

\citet{chen2020dasnet} designed a dual-attention Siamese network (DASNet) with double-margin contrastive loss that relieves the effect of pseudo-changes and captures long-range dependencies for robust discriminant feature extraction. \citet{lv2020object} devised an object key-point distance measurement module that improves the robustness of a model and relieves the effect of pseudo-change caused by the spectral domain difference of images. \citet{zhang2021foda} further reduced the impact of domain differences between bi-temporal images by using an adversarial learning procedure to align the features extracted from paired images. This reduces the negative effect of background variables and improves the accuracy of recognized building changes. Additionally, \citet{zhang2021object} developed an instance-level dual correlation attention-guided detector to solve the problem of pseudo-change and showed that this detector outperforms pixel-level-based contrastive learning approaches.

\citet{chen2021adversarial} devised instance-level change augmentation, which solves unbalanced change and label shortage problems by simulating samples via a generative adversarial network (GAN) to augment instance-level labels \citep{goodfellow2020generative} and achieves accuracy comparable to that of other methods but from significantly fewer samples. \citet{peng2020semicdnet} constructed a semisupervised CNN framework based on a GAN for change detection and used it to train models with partially labeled samples, which they showed decreased the demands of annotating labels and increased the ability of generalization. In addition, \citet{dong2021multiscale} developed a multi-scale context aggregation network to obtain detailed spatial information for unbalanced change in buildings, while \citet{li2020label} designed a robust noise-label active learning method to automatically screen reliable and informative positive samples.

\paragraph{3) Transfer learning}
Transfer learning \citep{pan2009survey} applies knowledge learned from sufficient labeled data to another task with insufficient samples to realize improved performance without the need for extensive labeling efforts. Generally, the changed buildings are rare compared with the backgrounds and unchanged buildings in pixels, especially over a short time interval. Therefore, to sufficiently exploit potential contextual information, many researchers have transferred knowledge of buildings to models via an auxiliary semantic segmentation branch to improve models' generalizability and thereby achieve robust extraction of building change.

\citet{sun2020fine} developed a multi-task learning strategy composed of change detection and an auxiliary building segmentation branch that makes full use of building semantics and realized improved change-detection performance. \citep{gao2022built,shen2022semantic} constructed a multi-task building change-detection network based on transfer learning that fully exploits prior information of building masks to optimize the extraction of building changes. 
Similarly, \citet{chen2022fccdn} devised a feature constraint change detection (FCCDN) network that consists of two encoders and an auxiliary segmentation branch and extracts discriminating building-change features. The LGPNet \citep{liu2021building} improved the robustness of change recognition models by combining local and global pyramid modules to capture discriminating features of changed buildings and enhanced the generalizability by employing a cross-task transfer learning strategy. \citet{liu2020building} developed a dual-task constrained Siamese network (DTCDSCN) that combines change detection with semantic segmentation and attention mechanisms to improve object-level discriminating feature representation.
\citet{li2022crossgeonet} proposed the CrossGeoNet to generate building footprints for label-scarce regions. \citet{kang2022disoptnet} and \citet{zheng2021deep} distilled important semantic knowledge from optical images into other data sources based on transfer learning for building segmentation.

\section{Method}
\label{s:methodology}
\subsection{Overview and problem setup}
The preliminary building information contained in historical building masks reveals a building's distribution at a previous time and thus can be used to validate building changes. Thus, unlike existing methods, which extract building changes from registered bi-temporal images and corresponding change labels, we designed and used multi-task segmentation decoders to extract newly constructed and removed buildings, respectively, by combining up-to-date images with building labels fully annotated with change states. Specifically, we designed the DCN-based feature transform model to alleviate the misidentification caused by the various misalignment between the historical footprints and up-to-date images. Furthermore, we proposed an instance-level contrastive loss to focus more on the global discrimination between buildings and backgrounds for robustness building change extraction. At last, considering the imbalanced change labels, we effectively generate simulated change samples by modifying the changing state of fully annotated labels randomly.

\subsubsection{Architecture of BCE-Net}

\begin{figure}[H]
	\centering
	\includegraphics[width=0.7\linewidth]{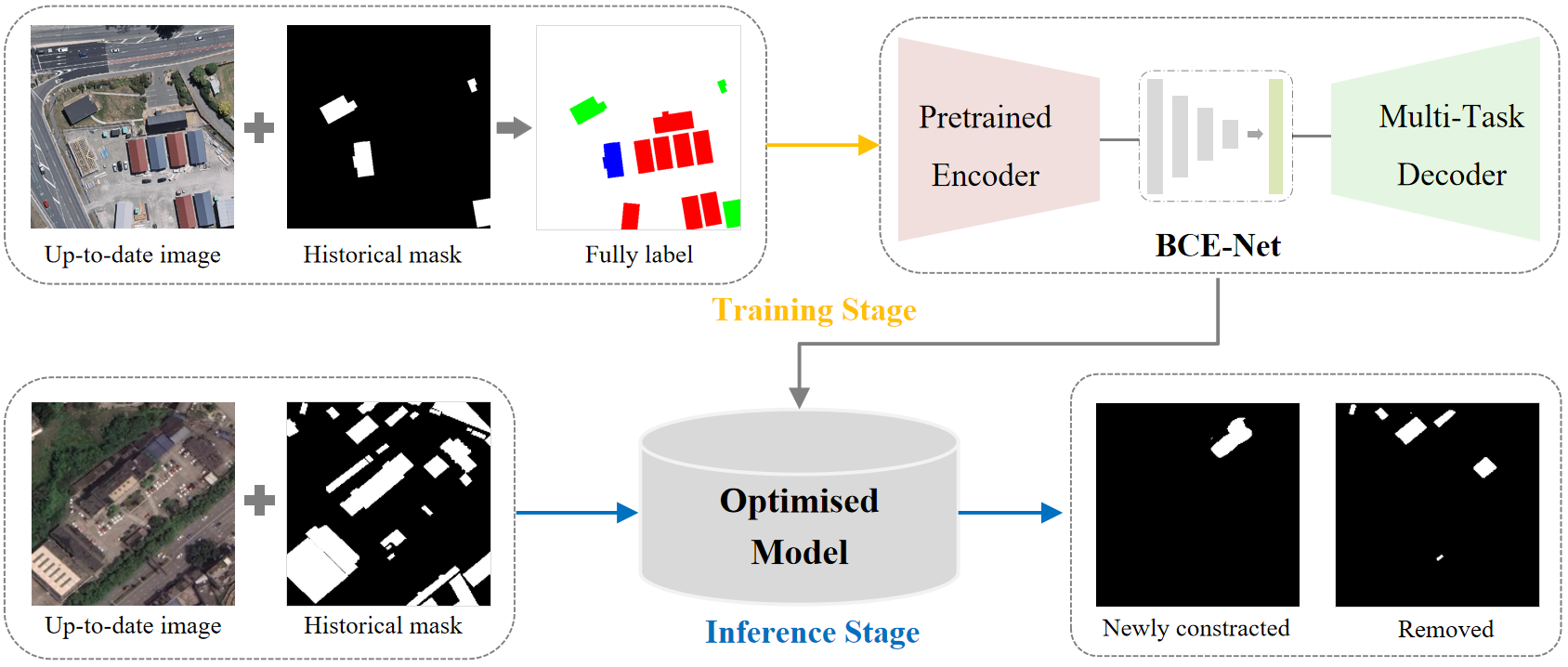}
	\caption{Overview of our framework for building change extraction. During the training stage, the up-to-date images are combined with fully annotated labels of building change states and corresponding historical building masks are used as inputs. During the inference stage, the model simultaneously extracts newly constructed and removed buildings with reference to images and matched historical building footprints.}
	\label{fig:fig3}
\end{figure}

An overview of our framework is given in Figure \ref{fig:fig3}. The fully annotated labels colored newly constructed, removed, and unchanged buildings in red, blue, and green, respectively, with these labels determined by combining images and historical building masks. Historical building masks are actually a subset of fully annotated change labels, which include those for all of the unchanged buildings and removed buildings. Our objective was to simultaneously extract newly constructed and removed buildings by combining single temporal up-to-date imagery and historical building information during the reference stage, thereby providing a route towards updating existing building databases.

\begin{figure}[H]
	\centering
	\includegraphics[width=0.9\linewidth]{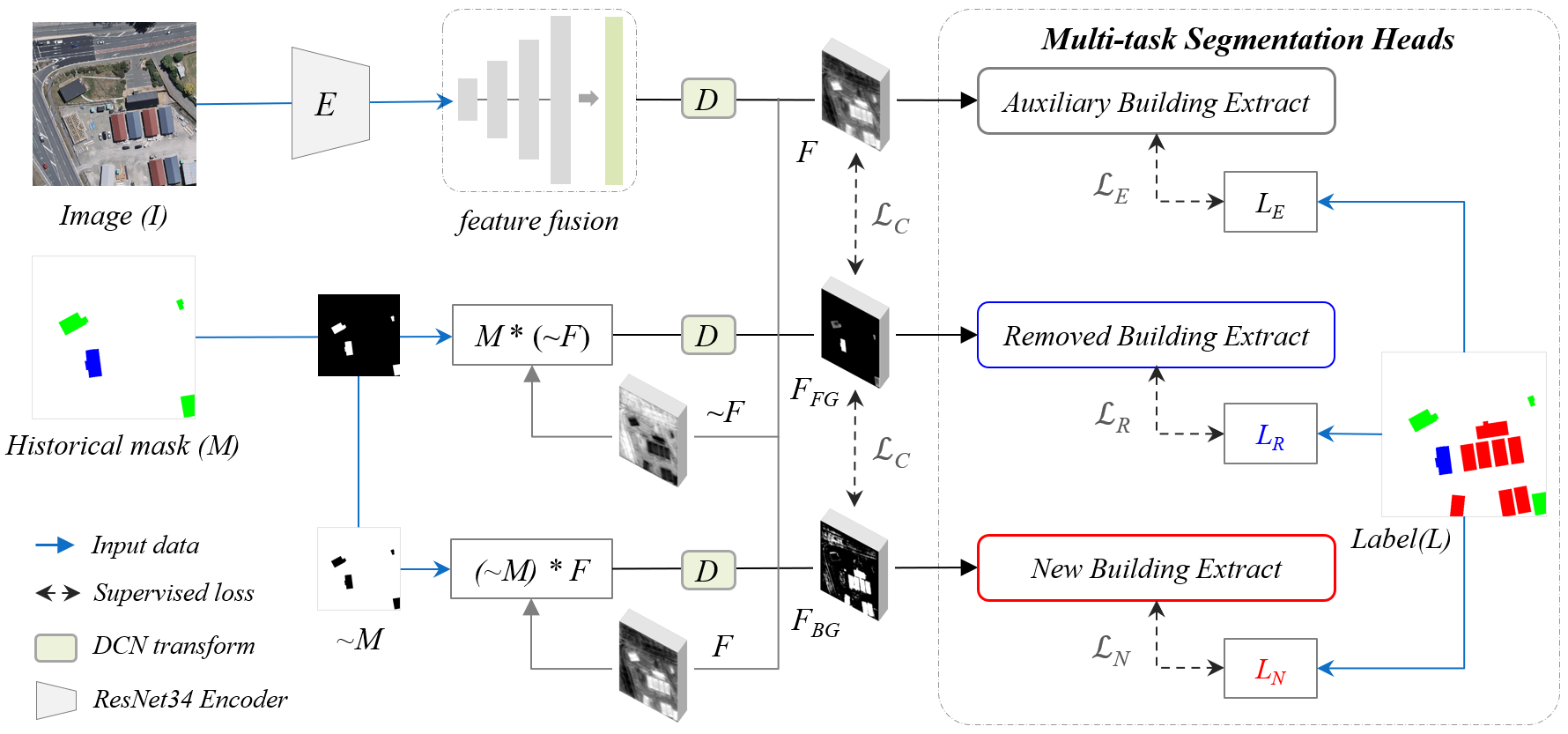}
	\caption{Overview of BCE-Net, which is composed of the following three modules: a pre-trained multi-level feature extractor; a DCN-based feature transformation module; and multi-task segmentation branches for extraction of newly constructed, removed, and existing buildings.}
	\label{fig:fig4}
\end{figure}

BCE-Net consists of four parts: a pre-trained encoder for extracting robust multi-level features; multi-task segmentation branches for extraction of newly constructed, removed, and existing buildings; a DCN-based transform module for consistent adaptive adjustment of features; and a building instance-constrained contrastive learning module for discriminating feature optimization. The architecture of BCE-Net is depicted in Figure \ref{fig:fig4} and its use is detailed in the following paragraph.

First, BCE-Net extracts multi-scale features through a pre-trained ResNet34\footnote{https://download.pytorch.org/models/resnet34-b627a593.pth} encoder and fuses them by up-sampling and concatenating addition to obtain features with one-quarter of the resolution of the input image. Second, BCE-Net splits the fused features into foreground features and background features (according to the historical building masks) for multi-task segmentation. Given the offset between historical masks and feature representation, we designed a DCN-based transform module to adjust features adjacent to historical building footprints. Third, the newly constructed, removed, and existing buildings are simultaneously extracted through a designed multi-task segmentation head. Finally, self-designed loss functions $\mathcal{L}_{N}$, $\mathcal{L}_{R}$, and $\mathcal{L}_{E}$ are used to optimize the extracted results with respect to the corresponding supervision labels $L_{N}$, $L_{R}$, and $L_{E}$, respectively. In addition, a self-designed instance-level contrastive loss constrained by changed building polygons is used to learn discriminating building representations based on global similarities and differences between features extracted from each branch.

\subsubsection{Problem setup}

Formally, the inputs of the model consist of the up-to-date image $I$ and the fully annotated building label $L$ – which comprises the newly constructed buildings $L_{N}$ (colored red), removed buildings $L_{R}$ (colored blue), and unchanged buildings (colored green) – obtained by combining $I$ and a historical building mask $M$. The multi-scale features are extracted by encoder $E$ and fused via channel concatenation and attention enhancement to form $F$. Additionally, $\sim F$ and $\sim M$ represent the reversed $F$ and $M$, respectively, which subtract themselves from the all-one matrix $\mathbb J$.

We designed three segmentation heads for newly constructed, removed, and existing building extraction, represented as $H_{N}$, $H_{R}$, and $H_{E}$, respectively. On the basis of $F$, background features $F_{BG}$ are split as the input of $H_{N}$ to predict newly constructed buildings $P_{N}$ with the supervised loss $\mathcal{L}_{N}$. Similarly, the foreground features $F_{FG}$ are split as the input of $H_{R}$ to predict removed buildings $P_{R}$ with the supervised loss $\mathcal{L}_{R}$. In addition, an auxiliary $H_{E}$ predicts existing buildings $P_{E}$ from $F$, with an optimized loss $\mathcal{L}_{E}$, to fully exploit building semantics. 

These pixel-wise losses joint the binary cross-entropy $\mathcal{L}_{B}$ and Dice loss $\mathcal{L}_{D}$ with $\alpha$ and $\beta$ weightings. Given the inconsistency between $M$ and $I$, which is caused by the offsets of building rooftops, deformable convolution layers are used to adaptively adjust the features before and after the split operation. To further enhance the features at a more global level, a self-designed similarity-based contrastive loss $\mathcal{L}_{C}$ is used to apply building-instance constraints to optimize the features in each branch.

\subsection{Multi-task decoders for building change extraction}
\label{s:multi_task_decoders}

\begin{figure}[H]
    \centering
    \includegraphics[width=0.8\textwidth]{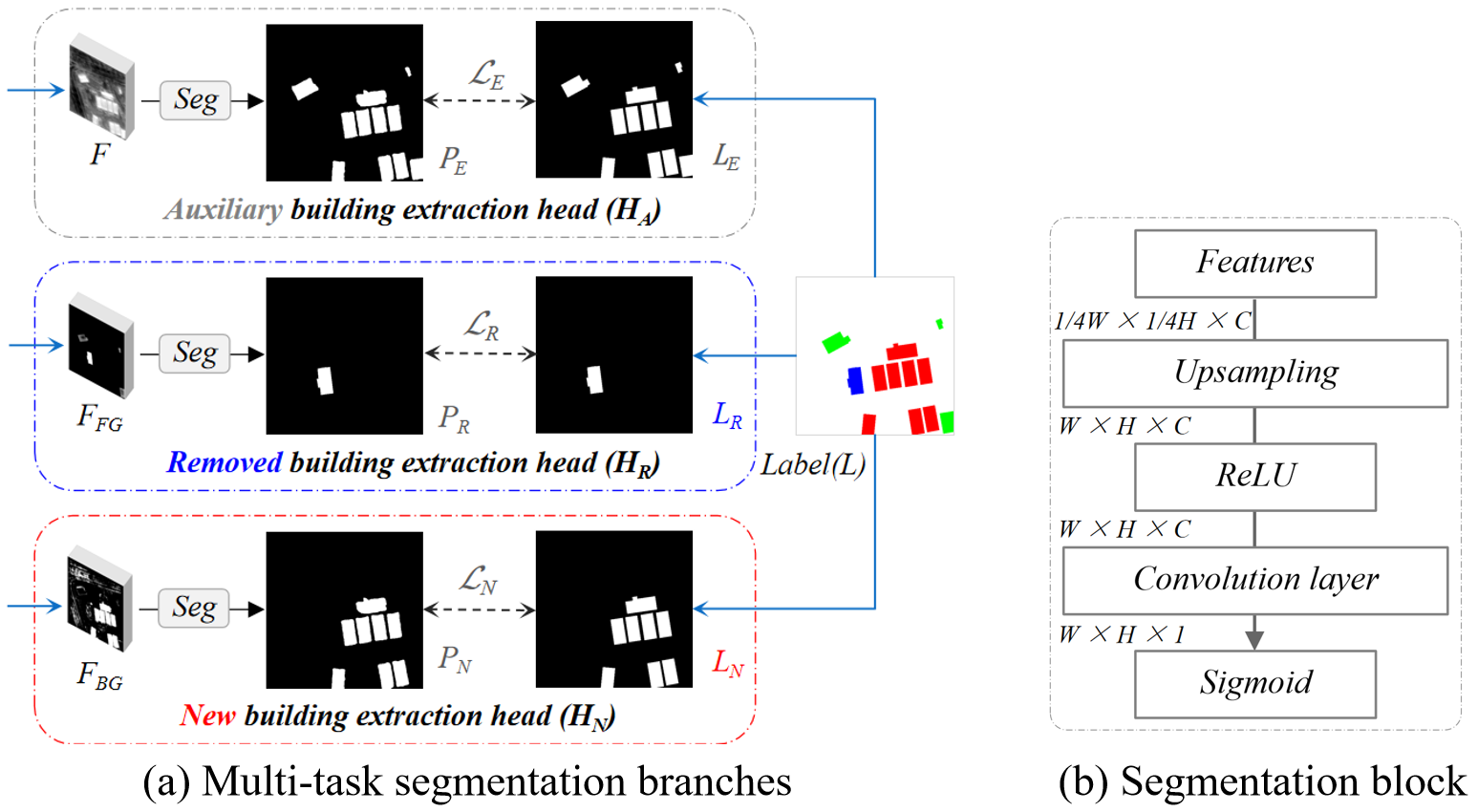}
    \caption{Detailed illustration of the multi-task segmentation branches.}
    \label{fig:fig5}
\end{figure}

BCE-Net extracts building changes, including newly constructed and removed buildings, via multiple segmentation tasks. In addition, we designed an auxiliary building segmentation branch that constrains extracted multi-scale features with potentially rich contextual semantics of existing buildings, as shown in Figure \ref{fig:fig5} (a). $M$ are composed of removed $L_{R}$ and unchanged buildings, while the existing buildings $L_{E}$ are composed of newly constructed $L_{N}$ and unchanged buildings. During the training and inference stage, the masks and ground-truth labels are converted into binary maps in which backgrounds and buildings are allocated the values 0 and 1, respectively. The detail of the segmentation block of three heads is shown in Figure \ref{fig:fig5} (b), which reveals that the block consists of a bilinear interpolation operator for upsampling the features to the same level as the input images and a convolution layer to reduce the channel to one, which is followed by a sigmoid activation function.

\subsubsection{Newly constructed building extraction}
For extracting newly constructed buildings, a spatial matrix multiplication is first performed to obtain the background features $F_{BG}$, which contain only the features $F$ not in spatial locations corresponding to the historical background masks, represented as $\sim M$. The $F_{BG}$ is calculated by Formula (\ref{eq:eq1}). Then, a segmentation head $H_{N}$ distinguishes the newly constructed buildings from $F_{BG}$.

\begin{equation}
	\begin{split}
	\sim\boldsymbol{M}&=\mathbb{J}-\boldsymbol{M},\\
	\boldsymbol{F_{BG}}&=(\sim\boldsymbol{M}) \times \boldsymbol{F}
	\end{split}
	\label{eq:eq1}
\end{equation}

During the training stage, the predicted pixel-wise newly constructed buildings $P_{N}$ $\in$ [0,1] represent the probability of a pixel belonging to a building. The annotated newly constructed labels $L_{N}$ supervise the prediction $P_{N}$ through a joint loss function $\mathcal{L}_{N}$, which includes a binary cross-entropy $\mathcal{L}_{B}$ and Dice loss $\mathcal{L}_{D}$, as given in Formula (\ref{eq:eq2}). We improved the model convergence by achieving a balanced optimization by introducing $\alpha$ and $\beta$ as weightings for the loss $\mathcal{L}_{B}$ and $\mathcal{L}_{D}$, which is motivated by \citep{guo2018dynamic,kendall2018multi}.

\begin{equation}	
	\mathcal{L}_{N}=\alpha*\mathcal{L}_{B} (P_{N},L_{N})+ \beta *\mathcal{L}_{D}(P_{N},L_{N})
	\label{eq:eq2}
\end{equation}

\subsubsection{Removed building extraction}
For extracting removed buildings, a spatial matrix multiplication is first performed to obtain the foreground features $F_{FG}$ by regarding fused features $F$ as corresponding to the historical building masks $M$ in space. Due to the contrary feature representation of removed and existing buildings, $F$ is reversed to $\sim{F}$ before the spatial splitting is conducted. As a result, a uniform segmentation head $H_{R}$ is used to identify a removed building. The $F_{FG}$ is calculated using Formula (\ref{eq:eq3}), with a sigmoid function $S$ applied to normalize $F$ before it is reversed.

\begin{equation}
	\begin{split}
		\sim\boldsymbol{F}&=\mathbb{J}-\boldsymbol{S}(\boldsymbol{F}),\\
		\boldsymbol{F_{FG}}&=\boldsymbol{M} \times (\sim\boldsymbol{F})
	\end{split}
	\label{eq:eq3}
\end{equation}

During the training stage, the predicted results $P_{R}$ $\in$ [0,1] represent the probability of belonging to the removed building. The annotated removed labels $L_{R}$ supervise the prediction $P_{R}$ through the joint loss function $\mathcal{L}_{R}$, which combines the binary cross-entropy $\mathcal{L}_{B}$ and Dice loss $\mathcal{L}_{D}$, as follows (Formula (\ref{eq:eq4})):

\begin{equation}	
	\mathcal{L}_{R}=\alpha*\mathcal{L}_{B} (P_{R},L_{R})+ \beta *\mathcal{L}_{D}(P_{R},L_{R})
	\label{eq:eq4}
\end{equation}

Generally, a removed building must be a specific polygon of historical building masks $M$. Thus, during the inference process, a specific historical building polygon is only identified as removed when the average probability of that polygon in the predicted $P_{R}$ is higher than a threshold $\theta$, which was 0.5 in this study. The identified removed buildings are also regularised, which directly supports updating of a historical building database.

\subsubsection{Auxiliary extraction of existing buildings}

The segmentation heads for extracting newly constructed buildings and removed buildings are only based on a partition of the features $F$. Moreover, only changed labels are used to supervise the models, so these are rarely compared with unchanged buildings. Therefore, to fully use potential semantic features of existing buildings and constrain the encoder to learn discriminating features for distinguishing buildings from backgrounds, we designed an auxiliary segmentation head $H_{E}$ to extract existing buildings corresponding to up-to-date images.

Similarly, during the training stage, the predicted pixel-wise results $P_{E}$ $\in$ [0,1] are supervised by the annotated existing building labels $L_{E}$, which consist of the newly constructed buildings and unchanged buildings and the joint loss function $\mathcal{L}_{E}$, which is a combination of $\mathcal{L}_{B}$ and $\mathcal{L}_{D}$ and is calculated as follows (Formula (\ref{eq:eq5})):

\begin{equation}	
	\mathcal{L}_{E}=\alpha*\mathcal{L}_{B} (P_{E},L_{E})+ \beta *\mathcal{L}_{D}(P_{E},L_{E})
	\label{eq:eq5}
\end{equation}

\subsection{Deformable convolution for feature adjustment}
\label{s:deformable_cnn}

As shown in Figure \ref{fig:fig1} (b), offsets between building footprints and rooftops cause obvious inconsistencies between historical building masks $M$ and buildings, which are manifested as offsets in rooftops and façades with respect to up-to-date images. This unavoidable inconsistency means that features split for foreground $F_{FG}$ and background $F_{BG}$ constrained by $M$ are not exactly as expected, i.e., the features that represent buildings are not strictly in the range of $M$. These wrongly allocated features arise near the boundaries of M and introduce severe noise into predicted results, which presents a substantial obstacle to accurate extraction of building changes.

As standard CNN samples input features on a fixed grid $G$, it is difficult to model geometric transformations from an irregular spatial range. In mathematical terms, each location $p_{0}$ in an output feature map $y$ is a sum of the grid sampled from input $x$ weighted by $w$, as expressed in Formula (\ref{eq:eq6}). \citet{dai2017deformable} devised deformable CNNs to enhance the transformation modeling capability of CNNs. These deformable CNNs contain additional parameters, which enable them to automatically learn the offsets for each spatial position. This allows them to capture long-range relationships instead of fixed sample grids and thereby achieve refined feature representation for geometrically deformed objects. In simple terms, an augmented grid with offsets was automatically learned, as defined by Formula (\ref{eq:eq6}). As such, the range of inconsistency between the $M$ and feature representation can be considered a geometric transformation.

\begin{equation}
	\begin{split}
		y(p_0) &= \sum_{p_n \in G} w(p_n) \cdot x(p_0+p_n),\\
		y(p_0) &= \sum_{p_n \in G} w(p_n) \cdot x(p_0+p_n+\Delta p_n)
	\end{split}
	\label{eq:eq6}
\end{equation}

To alleviate the offset between the masks and feature representations of images. we designed a DCN-based feature transform module to learn the geometric relationships between oblique buildings and masks for adaptive feature adjustment. Thus, a deformable convolution layer is used to automatically learn the offsets for each spatial location, according to the input features. This knowledge of offsets enables networks to adaptively determine long-range relationships by using irregular sample grids, especially at the adjacency of buildings and backgrounds. As a result, inconsistency is significantly reduced during model optimization.

\begin{figure}[H]
	\centering
	\includegraphics[width=0.9\textwidth]{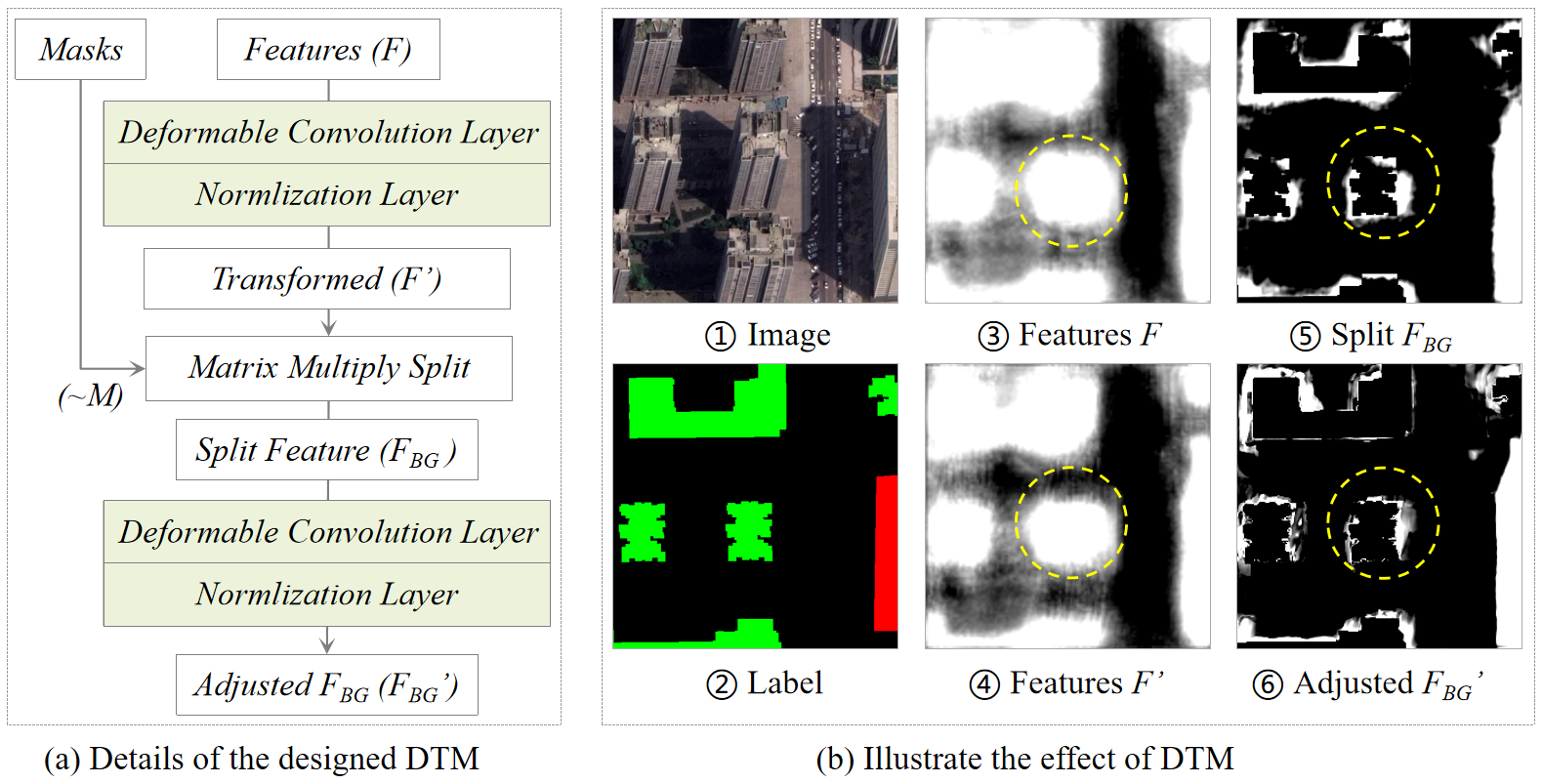}
	\caption{(a) Details of the DCN-based feature transformation module. (b) Illustration of the inconsistency correction performed by the DTM (indicated by the yellow circle).}
	\label{fig:fig6}
\end{figure}

The details of the designed DCN-based feature transform module (DTM) are shown in Figure \ref{fig:fig6} (a). First, the fused multi-scale features are transformed through a deformable convolution block (composed of a convolution layer followed by a normalization layer) to learn a global offset, supervised by the $L_{E}$ on the auxiliary building segmentation branch. Additionally, the split foreground $F_{FG}$ and background $F_{BG}$ are further adjusted through another deformable convolution block to correct the inconsistent spatial locations of features and masks. An example illustrating the effect of DTM is shown in Figure \ref{fig:fig6} (b). A comparison of \ding{174} with \ding{175} and \ding{176} with \ding{177}, each of which represents the features before and after transforming, respectively, reveals that the features were significantly optimized for adapting to the historical building masks. Note that, the $^{'}$ indicates features converted by the DTM that are not distinguished here, for brevity.

\subsection{Building instance constrained contrastive loss}

\begin{figure}[H]
	\centering
	\includegraphics[width=0.45\textwidth]{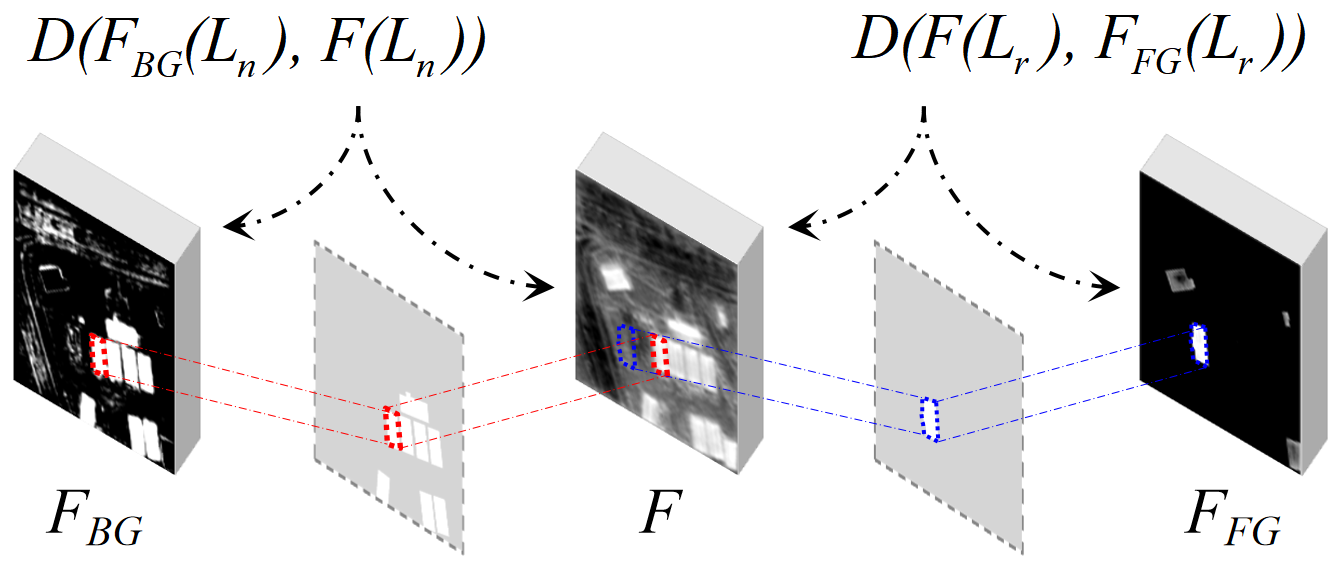}
	\caption{Illustration of building instance-constrained contrastive loss.}
	\label{fig:fig7}
\end{figure}

Pixel-wise contrastive learning has been widely used to constrain the distance of features among positive and negative samples \citep{chen2020dasnet,chen2020simple,khosla2020supervised}. However, although pixel-wise contrastive learning enhances the differences of features resulting from changes, it cannot avoid the fatal effects of pseudo-changes caused by backgrounds. To alleviate this problem, we constructed an instance-level contrastive loss $\mathcal{L}_{C}$ based on global feature representation constrained by building masks. This improves the discrimination of changed building features among the three segmentation branches and decreases the impact of pixel-level-pseudo changes. The designed instance-level contrastive loss is illustrated in Figure \ref{fig:fig7}.

As described in section \ref{s:deformable_cnn}, $F$, $F_{FG}$, and $F_{BG}$ are features of existing, removed, and newly constructed buildings, respectively, that are extracted from three segmentation heads. Similar to pixel-wise contrastive loss, the mappings of $F$ and $F_{BG}$ to newly constructed buildings should be close, while the mappings of $F$ and $F_{FG}$ to removed buildings should be more distant. Thus, cosine similarity is used to measure the distance $D$ of the features, where $D\in$ [-1,1]. This means that for each bounding box of newly constructed buildings that are labeled, the feature patches obtained from $F_{BG}$ and $F$ are represented as $F_{BG}^n$ and $F^n$, respectively, where $n\in$ [1,$N$] and $N$ represents the number of newly constructed buildings. Similarly, for each bounding box of removed buildings that are labeled, the features obtained from $F$ and $F_{FG}$ are represented as $F^r$ and $F_{FG}^r$, respectively, where $r\in$ [1,$R$] and $R$ represents the number of removed buildings. Before measuring the distance of features $F$, $F_{FG}$, and $F_{BG}$, a 1 × 1 convolution layer with normalization followed by a sigmoid activation function are used to generate consistent one-channel features and thus achieve stable model convergence. The instance-constrained contrastive loss $\mathcal{L}_{C}$ is calculated by Formula (\ref{eq:eq7}), as follows:

\begin{equation}
	\mathcal{L}_{C}=\dfrac{1}{2N}\sum_{n=1}^{N} (1-D(F_{BG}^{n},F^{n})) + \dfrac{1}{2R}\sum_{r=1}^{R} (1+D(F^{r},F_{FG}^{r}))
	\label{eq:eq7}
\end{equation}

\begin{figure}[H]
	\centering
	\includegraphics[width=0.72\textwidth]{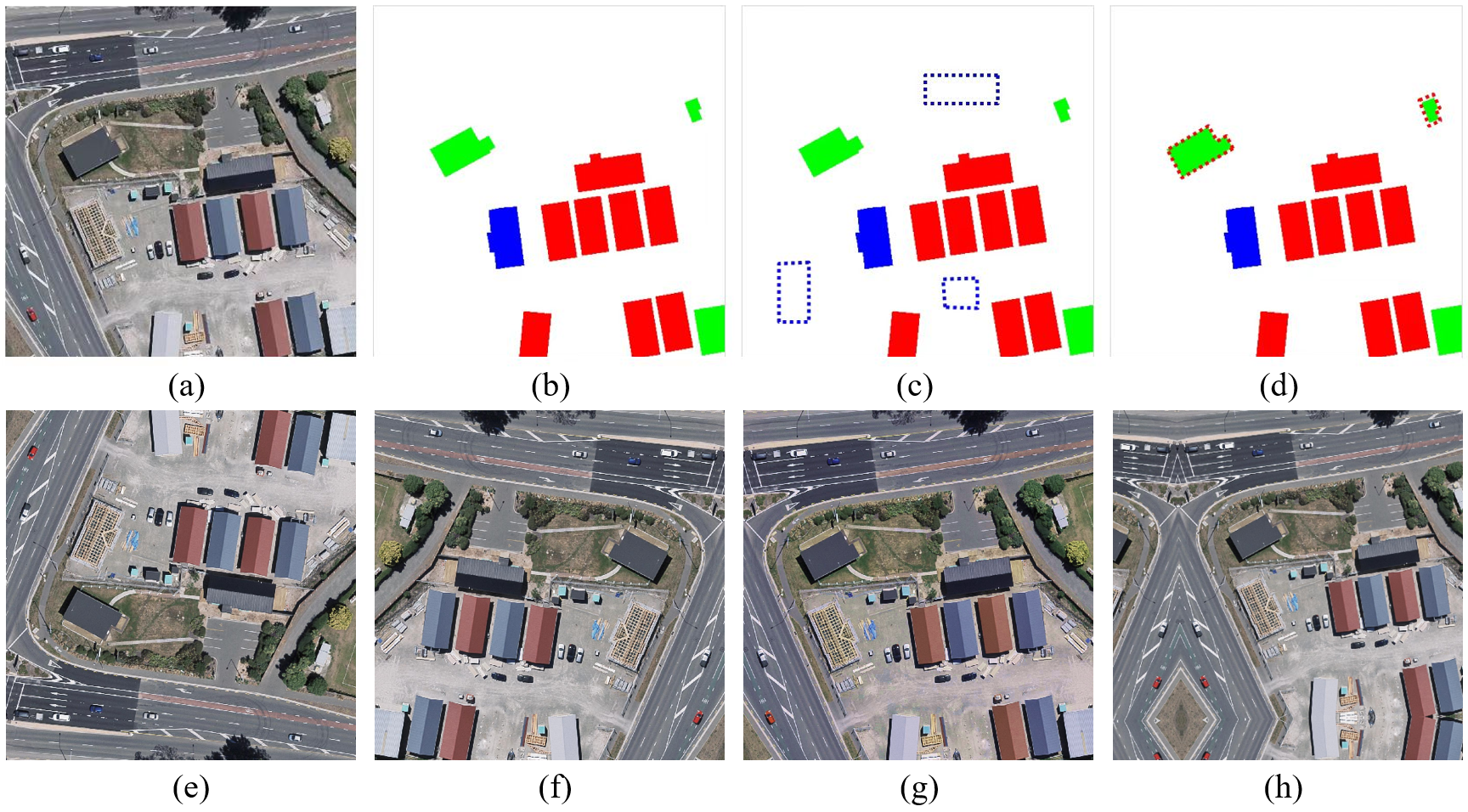}
	\caption{Example of change-label random generation and data augmentation strategy. (a) Image and (b) corresponding label. (c) Random generation of removed building labels. (d) Modification accounting for newly constructed buildings based on unchanged building labels. (e)-(h) Represent a sample augmented through vertical flipping, horizontal flipping, random color transformation, and random rotation and scaling, respectively.}
	\label{fig:fig8}
\end{figure}

The instance-level contrastive loss improves the ability of the method to extract discriminating features with a global constraint. However, this improvement is limited by the unbalanced labeling of changed buildings, as there are far fewer newly constructed and removed buildings than unchanged buildings. Thus, to further enhance performance, we developed a random sample generation strategy, which efficiently generates changed building labels based on existing building masks during the training stage, as illustrated in Figure \ref{fig:fig8}. Specifically, this approach simulates removed building masks with a specific regular area and angle on backgrounds, without overlapping with existing building labels, and then randomly marks unchanged buildings as newly constructed buildings. Crucially, the computational cost of this label-based sample augmentation is nearly negligible. Moreover, the generated change labels are realistic, unlike those generated by the popular GAN-based method that generates simulated buildings on images. We discuss the significance of the random sample-generation module in the ablation experiment in section 4.4.3. The total loss is the sum of the above four losses, which is calculated as follows (Formula (\ref{eq:eq8})).

\begin{equation}	
	\mathcal{L}=\mathcal{L}_{N} + \mathcal{L}_{R} +\mathcal{L}_{E} + \mathcal{L}_{C}
	\label{eq:eq8}
\end{equation}

\subsection{Evaluation metrics and strategies}

BCE-Net extracts building change masks from single temporal remotely sensed images via a classical binary semantic segmentation framework. Therefore, we quantitatively evaluated the extracted results with standard pixel-level segmentation metrics: precision, recall, F1 score, and intersection over union (IoU).
 
Our experiments were implemented using PyTorch and conducted on a single 2080Ti GPU. Stochastic gradient descent was used to optimize the models with an initial learning rate of 0.01. Our model was based on the pre-trained ResNet34. During the training stage, a data augmentation strategy was adopted that involved random flipping, rotation, random scaling, and color enhancement, as shown in the samples in the second column of Figure \ref{fig:fig8}. The models were trained for up to 100 epochs and the model that gave the best evaluation results was preserved.

\section{Experimental evaluation and analysis}
\label{s:experiments}
\subsection{Description of datasets}
\label{s:datasets}

To evaluate the significance of BCE-Net, we constructed the SI-BU dataset, which comprises single up-to-date images and fully annotated building labels for updating building changes. We obtained a fair comparison of the performance of our method with that of other methods by conducting comparative experiments using the public WHU-CD dataset \citep{ji2018fully}, as this contains building masks for two periods of images that can conveniently be converted into fully annotated change labels. The details of the two datasets are given below.

\paragraph{1) SI-BU dataset}
The SI-BU dataset comprises post-phase satellite imagery captured from Google Earth (Google Inc.) in 2021 of Guiyang, Guizhou province, China, along with corresponding labels. These labels were meticulously annotated by image interpretation experts to indicate the changes between the images and building masks collected from the same location in 2019. The dataset covers approximately 172 km$^2$ at a resolution of 0.5-0.8 m and contains areas with different levels of urbanization and development, with buildings significantly varying in height, scale, and appearance. In addition, the dataset exhibits an off-nadir problem, especially for high-rise buildings, due to the offset between building rooftops and footprints. This increases the challenge of automatically extracting building changes from the SI-BU dataset.

\begin{figure}[H]
	\centering
	\includegraphics[width=0.65\textwidth]{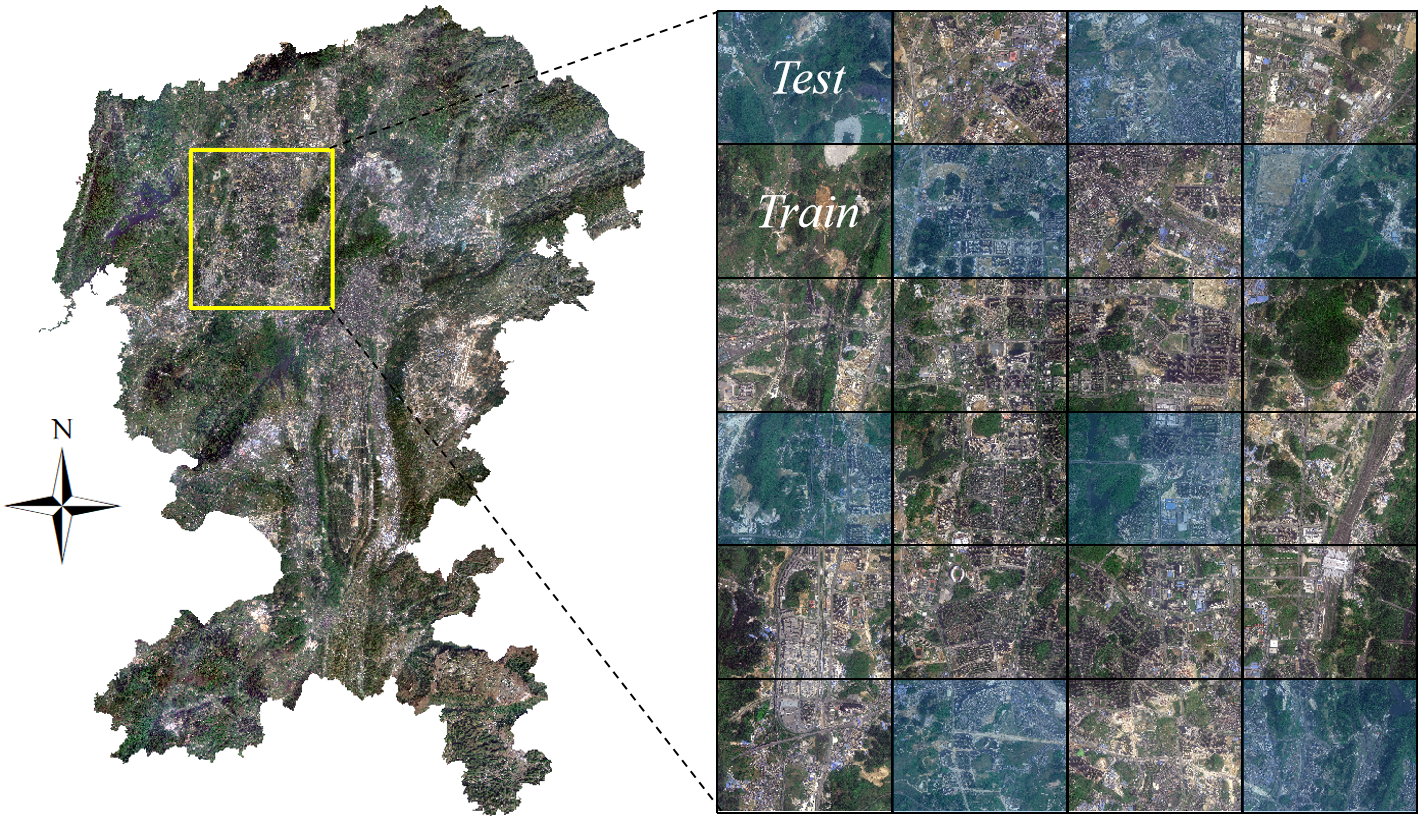}
	\caption{Overview of the SI-BU dataset, with the distribution of the test subsets colored light blue.}
	\label{fig:fig9}
\end{figure}
An overview of the SI-BU dataset is shown in Figure \ref{fig:fig9}. Given the location distribution and building density, 8 of 24 standard tiles were selected for testing, and the remaining standard tiles were used for training, to guarantee the independence of the training and test sets. The images and corresponding labels were cropped into 4,932 non-overlapping pairs, each with a size of 512 × 512 pixels, comprising 3,604 pairs for training and 1,328 pairs for testing. There are four labels, for background, unchanged buildings, newly constructed buildings, and removed buildings, which were allocated the values 0, 1, 2, and 3, respectively. Examples of labels are illustrated in Figure 2 (b), with the aforementioned values represented by the colors white, green, red, and blue, respectively.

\begin{figure}[H]
	\centering
	\includegraphics[width=0.8\textwidth]{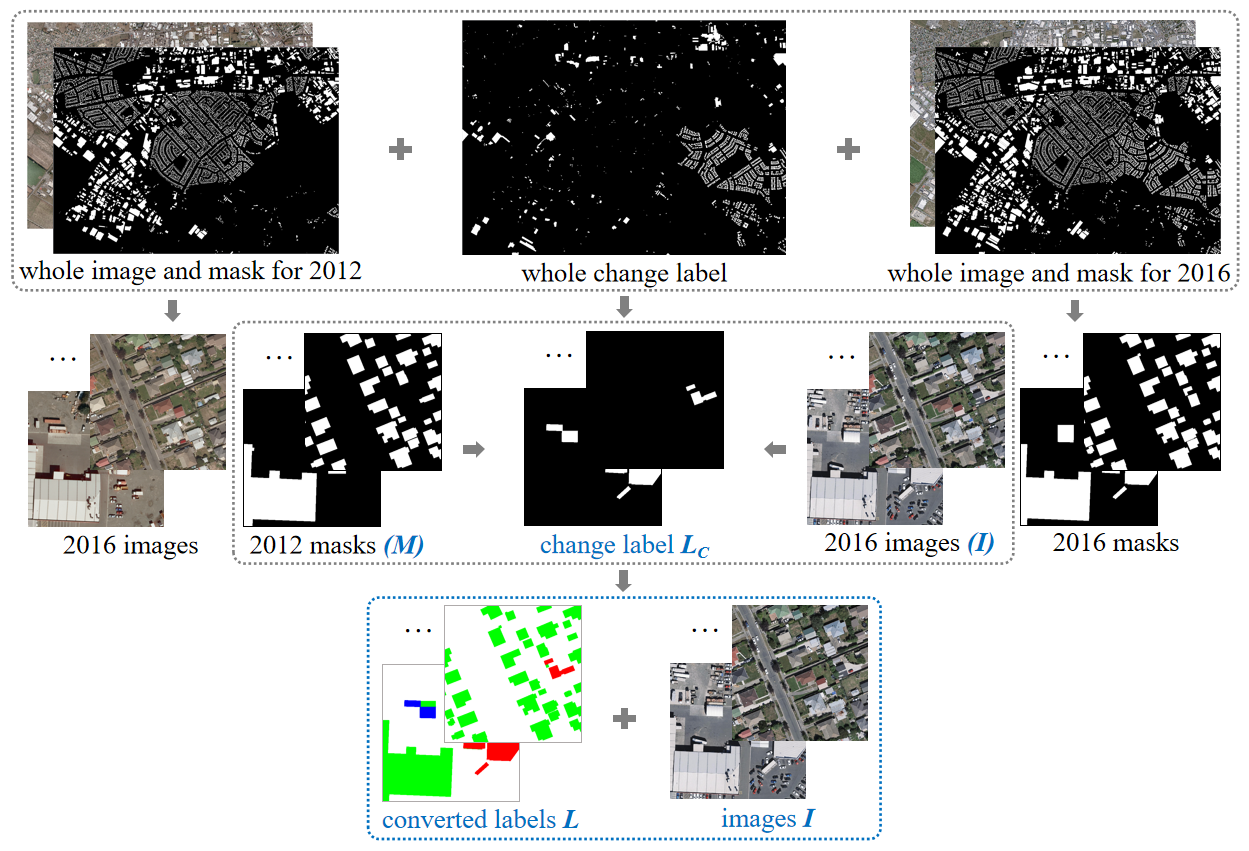}
	\caption{ Details of pre-processing and translation of the WHU dataset.}
	\label{fig:fig10}
\end{figure}

\paragraph{2) WHU-CD dataset}
The WHU-CD dataset consists of bi-temporal aerial images obtained in 2012 and 2016, and the corresponding full building masks and change labels at a resolution of 0.3 m. It provides 1,260 and 690 sliced tiles with a size of 512 × 512 pixels for images and corresponding building labels for training and testing, respectively, but it only provides the entire changed label. Researchers have divided this dataset in different ways, making it hard to fairly compare the performance of our method with that of other methods. Therefore, we supplemented the sliced change labels according to the official division, to aid future researchers and enable a comparison of our method with other methods.
Specifically, we used the following three-step process to transform the WHU-CD dataset into a single temporal form based on the provided change labels. First, the images captured in 2016 were converted to single-temporal images $I$. Second, full building labels corresponding to the 2012 historical building mask $M$ were converted to initial label $L$. Third, for each polygon $p$ in changed labels $L_C$, $p$ was marked as a removed building in $L$ if $p$ belonged to $M$; otherwise, $p$ was marked as a newly constructed building in $L$. We illustrate the transformation workflow, for clarity, in Figure \ref{fig:fig10}.

\begin{table}[H]
	\centering
	\caption{Quantity of images and percentage of pixel-level change in buildings in SI-BU and WHU-CD. PR, PN, and PU are the percentage of removed, new, and unchanged buildings.}
	\label{tab:tab1}
	
	\begin{tabular}{@{}ccccccccc}
		\toprule
		\multicolumn{2}{c}{Dataset} & Source  & No. of tiles  & Size   & Resolution(m)  & PR(\%)  & PN(\%)  & PU(\%)  \\
		\midrule
		\multirow{2}{*}{SI-BU}  & train & Satellite & 3,604 & 512  & 0.5–0.8 & 0.62	& 1.71 & 13.61 \\
		&test	& Satellite	& 1,329	& 512 	& 0.5–0.8	& 0.43	& 1.63	& 15.23 \\
		\multirow{2}{*}{WHU-CD}  & train	 & Aerial	& 1,260	 & 512 	 & 0.3	  & 0.27	 & 4.44	 & 14.67\\
		&test	 & Aerial	& 690	  & 512 & 0.3	 & 0.14	 & 3.41	 & 14.90\\
		\bottomrule
	\end{tabular}
\end{table}

As the SI-BU dataset was captured over only 2 years, it has a more serious imbalanced change-label problem than the WHU-CD dataset. Thus, the SI-BU dataset is more challenging and thus more practical than the WHU-CD dataset for investigating building changes. These two processed datasets will be made publicly available for academic research. More comparisons of the two datasets are shown in Table \ref{tab:tab1}.

\subsection{Ablation experiment}

To evaluate the effect of each module in our method, we performed ablation experiments using the SI-BU and WHU-CD datasets. All of the ablation experiments had the same hyperparameters, such as batch size, learning rate policy, and maximum training epoch. The experimental results are shown in Table \ref{tab:tab2} and Table \ref{tab:tab3}, respectively.

The baseline was designed based on the pre-trained ResNet34 backbone, with three segmentation branches used to extract the newly constructed, removed, and existing buildings, respectively. Subsequently, we added common strategies – including data augmentation (DA) and attention-based feature enhancement (ATT) – to compare performance improvements. Then, we gradually introduced a random sample generator (RSG), an instance-level contrastive learning (ICL) strategy, and a DCN-based feature transform module (DTM) to evaluate their contribution to model performance.

\begin{table}[H]
	\centering
	\caption{Results of the ablation experiment using the SI-BU dataset. This evaluated a data augmentation (DA) strategy, an attention-based enhancement (ATT) strategy, a random sample generator (RSG), an instance-level contrastive learning (ICL) strategy, and the DCN-based feature transform module (DTM)}
	\label{tab:tab2}
	\begin{tabular}{@{}@{}ccc|ccc|cccc}
		\toprule
		\multicolumn{6}{c}{Ablation Module} & \multicolumn{4}{c}{SI-BU} \\
		Baseline  & DA  & ATT  & RSG  & ICL  & DTM  & IoU &$\Delta IoU \uparrow$ & F1 &$\Delta F1 \uparrow$ \\ 
		\midrule
		\usym{1F5F8} & & & & & & 44.96	& -	& 62.03	& - \\
		\usym{1F5F8} & \usym{1F5F8}  & & & & & 50.08	 & +5.12	 & 66.74	 & +4.71 \\
		\usym{1F5F8} & \usym{1F5F8}  & \usym{1F5F8}  & & & & 50.36	 & +0.28	 & 66.99	 & +0.25 \\
		\usym{1F5F8} & \usym{1F5F8}  & \usym{1F5F8}  & \usym{1F5F8}  & & & 51.08	 & +0.72	 & 67.62	 & +0.63 \\
		\usym{1F5F8} & \usym{1F5F8}  & \usym{1F5F8}  & \usym{1F5F8}  & \usym{1F5F8}  & & 51.93	 & +0.85	 & 68.36	 & +0.74\\
		\usym{1F5F8} & \usym{1F5F8}  & \usym{1F5F8} & \usym{1F5F8} & \usym{1F5F8} & \usym{1F5F8}	& 54.73	 & +2.80	 & 70.74	 & +2.38 \\
		\bottomrule
	\end{tabular}
\end{table}

\begin{figure}[H]
	\centering
	\includegraphics[width=0.85\textwidth]{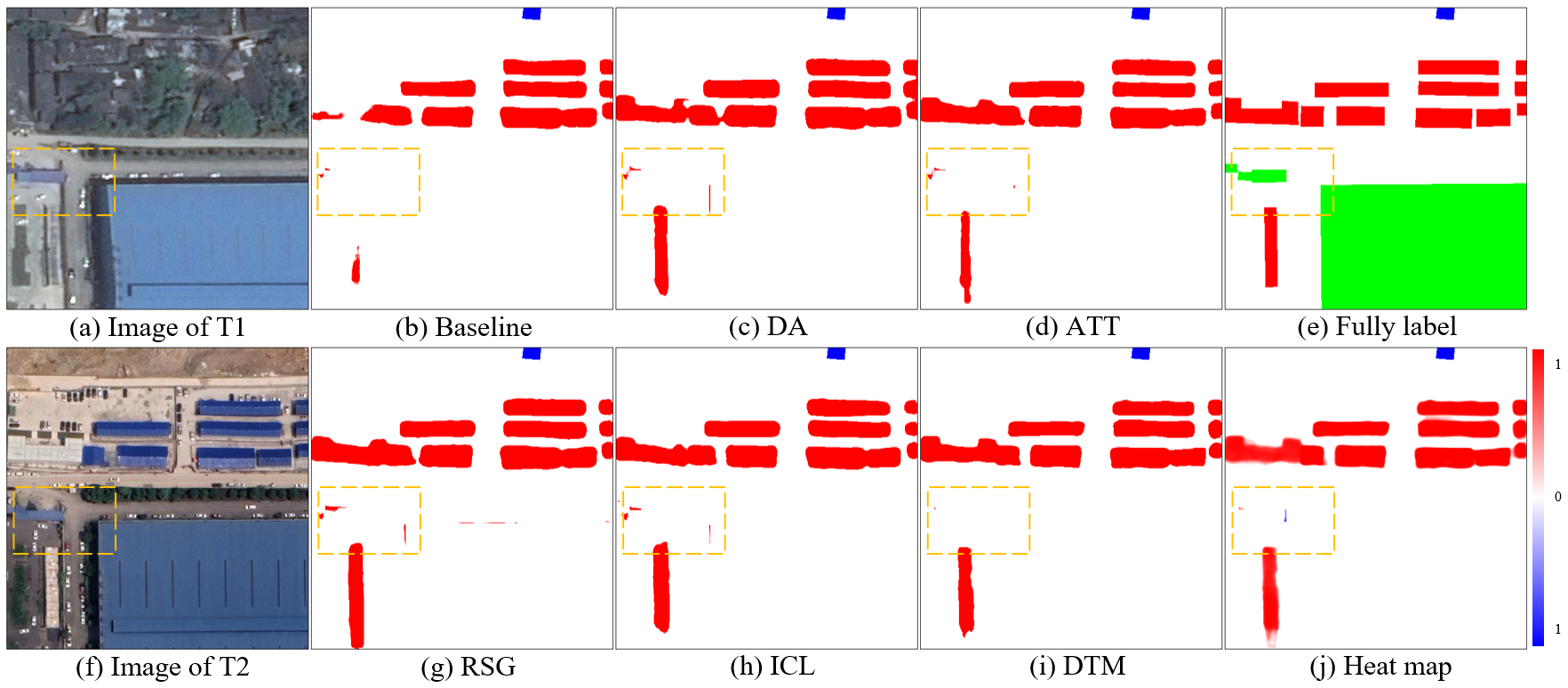}
	\caption{Results from ablation experiment using the SI-BU test set. (a) and (f) are images of the same area taken at different times; (b) is the baseline; (c), (d), (g), (h), and (i) are the results obtained after successively adding the DA, ATT, RSG, ICL, and DTM modules, respectively; (e) shows the buildings labeled with different colors according to their change state, where the red, blue, and green indicates new, removed, and unchanged buildings, respectively; and (j) represents the predicted probability maps corresponding to (i).}
	\label{fig:fig11}
\end{figure}

As shown in Table \ref{tab:tab2}, the baseline achieved an F1 score of 62.03\%. The commonly used DA and ATT strategies provided a 4.71 and 0.25 increase in F1 score, respectively. The strategies involving DA provided a significant increase in F1 score on this lower baseline, which contained more practical and challenging data. The RSG relieved the problem of unbalanced change labels and provided a 0.63 increase in F1 score. The ICL strategy provided a 0.74 increase in F1 score by fully utilizing the potential global semantics and learning the latent consistency of building features. The DTM provided a 2.38 increase in F1 score by adjusting and correcting the feature representation near the historical building footprints. Aside from the performance of the commonly used DA and ATT strategies, the performance increased by 4.37 in IoU and 3.75 in F1 score. We illustrate the predicted results obtained using the SI-BU dataset in Figure \ref{fig:fig11}, to provide a qualitative comparison. A comparison of (h) with (i) shows that the noise near the historical building footprints was significantly alleviated, which highlights the importance of the DTM.

\begin{figure}[H]
	\centering
	\includegraphics[width=0.84\textwidth]{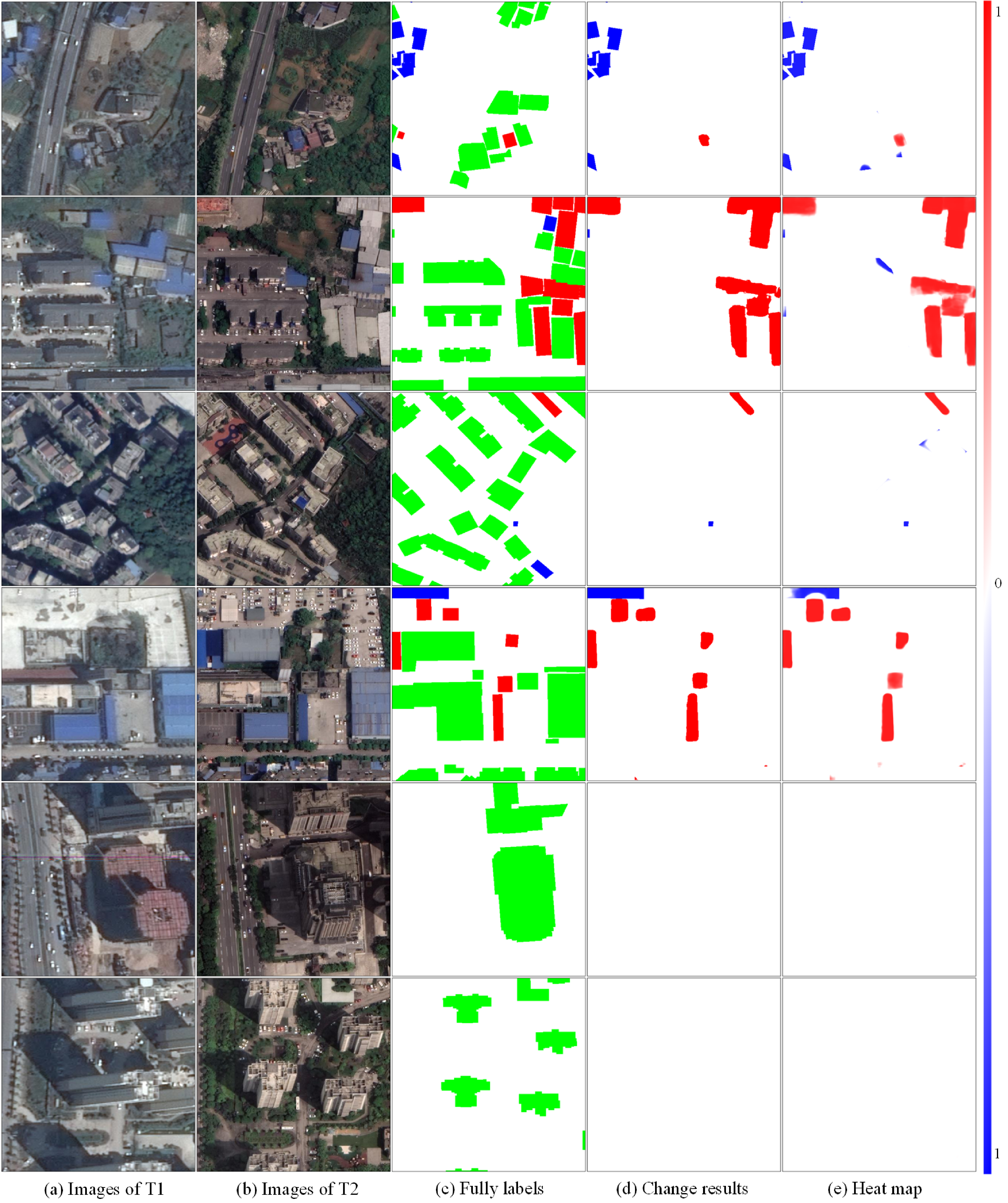}
	\caption{Building changes extracted using the SI-BU test dataset, where (a) represents historical images; (b), (c), and (d) contain up-to-date images, buildings fully labeled with their respective change state, and extracted building changes, respectively; and (e) is the predicted probability map corresponding to (d).}
	\label{fig:fig12}
\end{figure}

We provide more results obtained using the challenging SI-BU dataset in Figure \ref{fig:fig12} to illustrate the performance of BCE-Net. The first four rows illustrate that BCE-Net extracted the changed buildings, although the buildings exhibited an offset between the two data periods. Moreover, the samples in the last two rows reveal the challenge of the building off-nadir problem encountered when using approaches based on bi-temporal images, while demonstrating the robustness of BCE-Net to the significant offset between the two data periods.

\begin{table}[H]
	\centering
	\caption{Results obtained from the ablation experiment using the WHU-CD dataset. The evaluated modules were data augmentation (DA), attention-based enhancement (ATT), random sample generator module (RSG), instance-level contrastive learning strategy (ICL), and DCN-based feature transform (DTM) modules.}
	\label{tab:tab3}
	\begin{tabular}{@{}@{}ccc|ccc|cccc}
		\toprule
	 	\multicolumn{6}{c}{Ablation Module} & \multicolumn{4}{c}{WHU-CD} \\
	  	Baseline  & DA  & ATT  & RSG  & ICL  & DTM  & IoU & $\Delta IoU \uparrow$ & F1 &$\Delta F1 \uparrow$ \\ 
	 	\midrule
	 	\usym{1F5F8} & & & & & & 86.81	& -	& 92.94	& - \\
	 	\usym{1F5F8} & \usym{1F5F8}  & & & & & 87.11	& +0.30	& 93.11	& +0.17 \\
	 	\usym{1F5F8} & \usym{1F5F8}  & \usym{1F5F8}  & & & & 87.37	& +0.26	& 93.26	& +0.15 \\
	 	\usym{1F5F8} & \usym{1F5F8}  & \usym{1F5F8}  & \usym{1F5F8}  & & & 87.88	& +0.51	& 93.55	& +0.29 \\
	 	\usym{1F5F8} & \usym{1F5F8}  & \usym{1F5F8}  & \usym{1F5F8}  & \usym{1F5F8}  & & 88.34	& +0.46	& 93.81	& +0.26 \\
	 	\usym{1F5F8} & \usym{1F5F8}  & \usym{1F5F8} & \usym{1F5F8} & \usym{1F5F8} & \usym{1F5F8}	& 88.66	& +0.32	& 93.99	& +0.18 \\
		\bottomrule
	\end{tabular}
\end{table}

\begin{figure}[H]
	\centering
	\includegraphics[width=0.85\textwidth]{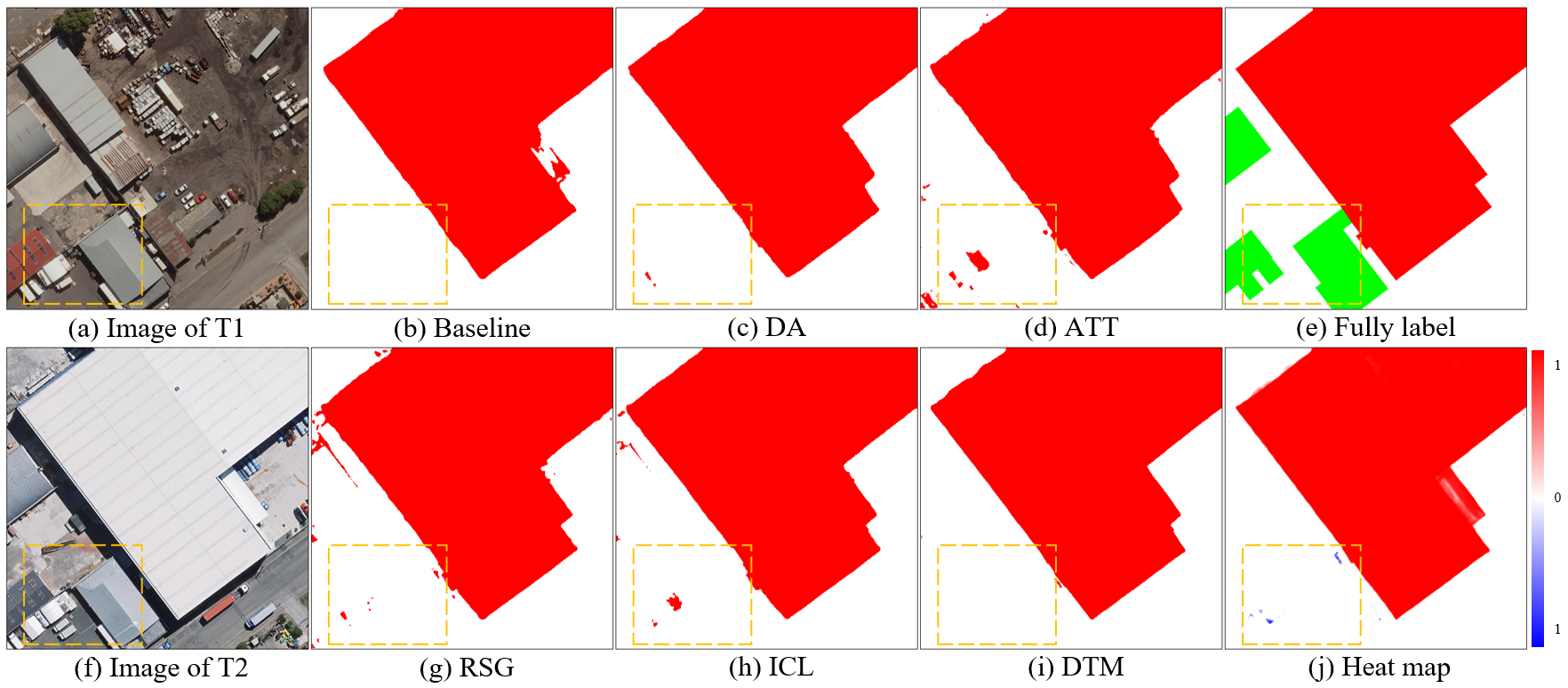}
	\caption{Results obtained from using the WHU-CD test set, where (a) and (f) are images of the same area taken at different times; (b) is the baseline; (c), (d), (g), (h), and (i) are the results obtained after gradually adding the DA, ATT, RSG, ICL, DTM modules, respectively; (e) shows buildings labeled with their respective change state, where the red, blue, and green colors indicate new, removed, and unchanged buildings, respectively; and (j) is the predicted probability map corresponding to (i).}
	\label{fig:fig13}
\end{figure}

In contrast, our baseline achieved an F1 score of 92.94\% using the WHU-CD test set, as shown in Table \ref{tab:tab3}. The commonly applied DA and ATT strategies provided 0.32 increases in F1 score; the RSG and ICL strategies provided 0.29 and 0.26 increases in F1 score, respectively; and the DTM provided a 0.18 increase in F1 score. The main contribution was a 1.29 increase in IoU and a 0.73 increase in F1 score. We illustrate the predicted results obtained from using the WHU-CD dataset in Figure \ref{fig:fig13}, to provide a qualitative comparison.

\subsection{Performance comparison}
BCE-Net is totally different from the majority of existing building change-detection methods, both in its theoretical basis and data demands. To fairly compare the building change-extraction performance of BCE-Net with that of other methods, we conducted a comparative experiment using the WHU-CD dataset.

Numerous related methods use the WHU-CD dataset but employ different data partitions. In this study, we compared the performance of BCE-Net with various existing methods that adopted a clear and dominant data division, including HFA-Net, SDACD \citep{liu2022end}, SRCDNet, DSA-Net, EGRCNN, DTCDSCN, DASNet, DDCNN, LGPNet, STPNet, SwinSUNet, SFCCD, FCCDN, and MTCDN. The SDACD uses a collaborative domain adaptation for images and features, which relieves the domain variation between bi-temporal images and the pseudo-change caused by environmental differences. The SwinSUNet method uses a pure Transformer network based on the Siamese framework to capture long-term global relationships and achieves state-of-the-art (SOTA) performance using the WHU-CD dataset. Notably, the SFCCD, FCCDN, and MTCDN approaches utilize building masks of two-period images as supplementary supervision, which is consistent with our approach in terms of data requirements.

These methods conduct experiments by partitioning data in two ways: the first is the official partitioning, and the second is a random 8:2 division for training and testing, respectively, with this division performed after clipping the entire dataset into patches with a resolution of 256 × 256 pixels. Thus, we conducted our experiments using each of these data partitions to enable a fair comparison, except that we converted the samples as described in section \ref{s:datasets}. The experimental results are shown in Table \ref{tab:tab4}. We compared our results with those reported by the authors for each of the other methods, where $\ast$ indicates methods that used an official data division.

\begin{table}[H]
	\centering
	\caption{Comparison of the performance of BCE-Net and related methods using the WHU-CD test dataset.}
	\label{tab:tab4}
	\begin{tabular}{l|cccc}
		\toprule
		Method  & Precision (\%) & Recall (\%) & F1 Score(\%) & IoU(\%) \\
		\midrule
		HFA-Net $\ast$ \citep{zheng2022hfa}	& --	  & --	    & 88.23	    & 78.93 \\
		SDACD $\ast$ \citep{liu2022end}	 	& 93.85	  & 90.91	& 92.36		& 85.80 \\
		BCE-Net $\ast$ (Ours)	 				& 95.28	  & 92.74	& 93.99	 	& 88.67 \\
		\midrule
		SRCDNet \citep{liu2021super}	 & 84.84	  & 90.13	   & 87.40	 	& 77.63 \\
		DSA-NET \citep{ding2021dsa}	 	 & 89.35	  & 87.63	   & 88.48	 	& 79.35 \\
		EGRCNN \citep{bai2021edge}	 	 & 89.90±0.98 & 87.95±1.24 & 88.90±0.43	& 80.04 \\
		DTCDSCN \citep{liu2020building}	 & --	 	  & 89.32±0.50 & 89.01±0.80	& 79.08±0.13 \\
		DASNet \citep{chen2020dasnet}	 & 90.00	  & 90.50	   & 91.00	 	& 82.23 \\
		DDCNN \citep{peng2020optical}	 & 93.71	  & 89.12	   & 91.36	 	& 84.09 \\
		LGPNet \citep{liu2021building}	 & 93.84	  & 90.59	   & 92.19	 	& 85.51 \\
		STPNet \citep{yang2021spatio}	 & 95.52	  & 92.15	   & 92.84	 	& 86.63 \\
		SwinSUNet \citep{zhang2022swinsunet} & 95.00  & 92.60	   & 93.80	    & 88.30 \\
		\midrule
		SFCCD $\dagger$	\citep{shen2022semantic}	 	 & 93.66	  & 85.68	   & 89.49	    & 80.98 \\
		FCCDN $\dagger$ \citep{chen2022fccdn}	 	 & 96.39	  & 91.24	   & 93.73	    & 88.20 \\
		MTCDN $\dagger$ \citep{gao2022built}	 	 & 94.57	  & 92.93	   & 93.74	    & 88.22 \\
		BCE-Net (Ours)					 & 95.28	  & 93.98	   & 94.63	    & 89.80 \\
		\bottomrule
	\end{tabular}
\end{table}

On the officially provided test set, BCE-Net achieved an F1 score of 93.99\%, which is 1.63 greater than that of the SDACD method, which had achieved the SOTA result in this test set. Notably, our baseline model outperformed SDACD by 0.58 in terms of F1 score. On the custom partitioned WHU-CD dataset, BCE-Net obtained the highest F1 score of 94.63\%, surpassing all other methods. Specifically, BCE-Net outperformed MTCDN and FCCDN by approximately 0.89 and 0.9 in F1 score, respectively. These methods utilize the latent building semantic through an auxiliary segmentation branch with fully annotated building labels (indicated with $\dagger$). Moreover, the F1 score of BCE-Net was 0.83 greater than that of SwinSUNet, which had achieved the SOTA result using the WHU-CD dataset and is a Transformer-based method that has been widely proven to outperform CNN networks.

\begin{figure}[H]
	\centering
	\includegraphics[width=0.84\textwidth]{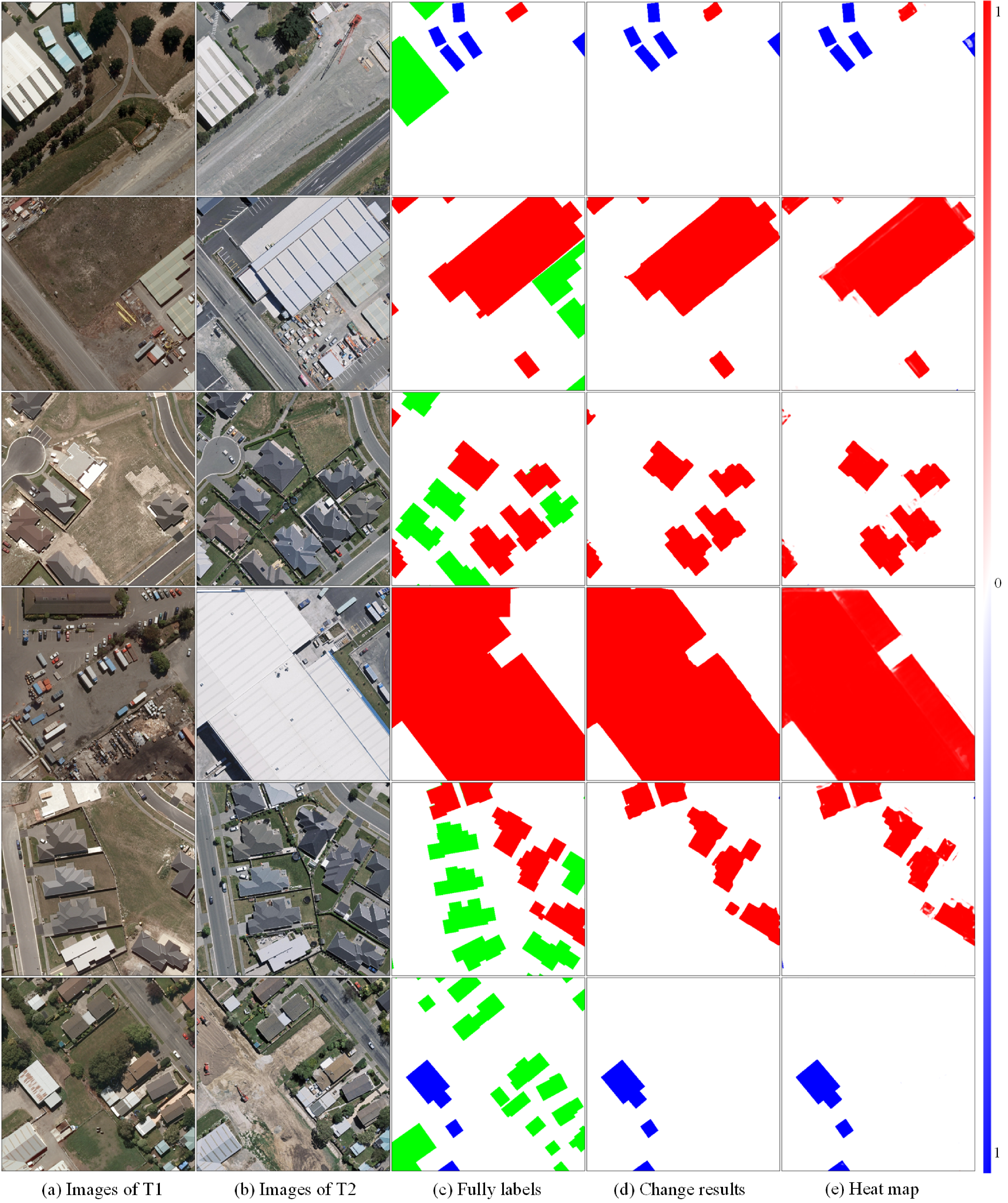}
	\caption{Samples of extracted changes in buildings determined using the WHU-CD test dataset, where (a) represents historical images; (b), (c), and (d) represent up-to-date images, buildings fully labeled with their respective change states, and extracted building changes, respectively; and (e) represents the predicted probability maps corresponding to (d).}
	\label{fig:fig14}
\end{figure}

The experimental results demonstrate the superiority of the multi-task framework of BCE-Net for extracting building changes by combining historical footprints and single temporal images. We illustrate more results obtained using the WHU-CD test in Figure \ref{fig:fig14} to provide a visual comparison.

\subsection{Efficiency analysis}
The ablation and comparison experiments performed using the SI-BU and WHU-CD datasets show that BCE-Net outperforms the other methods. Next, given the trade-off between precision and complexity, we evaluated the numbers of floating-point operations (FLOPs) and trainable parameters of the methods to compare their time and space complexity, respectively. Additionally, their IoU was calculated to compare the accuracy. In addition, for a fair comparison, we used an image with dimensions of 256 × 256 × 3 to calculate the complexity of other methods that have publicly available source code.

\begin{figure}[H]
	\centering
	\includegraphics[width=0.6\textwidth]{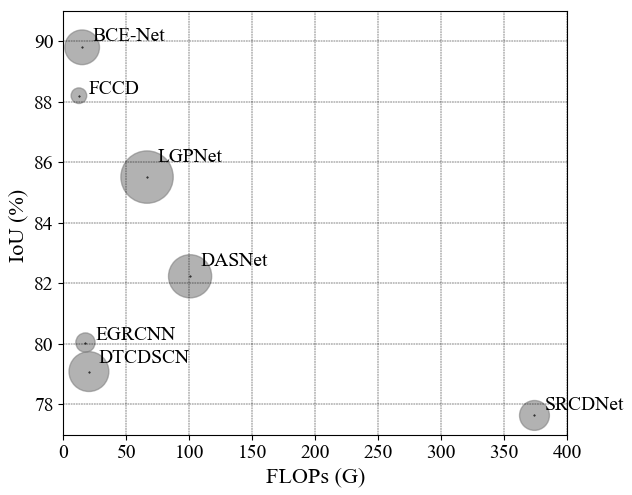}
	\caption{ Comparison of the complexity and accuracy of BCE-Net with those of other methods. The radius of a circle represents the number of trainable parameters.}
	\label{fig:fig15}
\end{figure}

The results are compared in Table \ref{tab:tab5}. BCE-Net is less complex than the other methods except for the FCCDN. Due to its three parallel multi-task decoders, BCE-Net has more trainable parameters than the EGRCNN and FCCDN, but BCE-Net has a similar number of FLOPs to EGRCNN and FCCDN. Moreover, BCE-Net has the best IoU. Thus, compared with the other methods, BCE-Net has superior accuracy and lower complexity: it achieves a better balance between accuracy and complexity. We illustrate the efficiency and accuracy of BCE-Net and other methods in Figure \ref{fig:fig15} to provide a visual comparison.

\begin{table}[H]
\centering
\caption{Comparison of the complexity of BCE-Net and other methods.}
\label{tab:tab5}
\begin{tabular}{lccc}
	\toprule
	Method  & FLOPs ($\times10^9$) & Parameters ($\times10^6$) & IoU(\%) \\
	\midrule
	SRCDNet	& 374.17	& 23.08		& 77.63 \\
	EGRCNN	& 17.64		& 9.63		& 80.04 \\
	DTCDSCN	& 20.39		& 41.07		& 79.08 \\
	DASNet	& 100.72	& 48.22		& 82.23 \\ 
	LGPNet	& 66.54		& 70.99		& 85.51 \\
	FCCDN	& 12.44		& 6.30		& 88.20 \\
	BCE-Net	& 15.00		& 31.16		& 89.80 \\
	\bottomrule
\end{tabular}
\end{table}

\subsection{Discussion and limitations}

The validation of buildings within a specific period according to up-to-date images to obtain information about newly constructed and removed buildings has great potential for application to routine building-change surveying and database updating. This is because reliable historical building masks can be obtained for specific periods from an existing building database or be generated automatically from imagery by a robust building segmentation model. Additionally, our novel approach can be used to identify changed buildings irrespective of the sequence of images and building masks.

Deep learning-based workflows for building change extraction require sufficient annotated samples. However, it is challenging to construct change labels manually from bi-temporal images with angle, illumination, and seasonal differences. In this study, we demonstrated a new approach for automatically constructing such samples from existing building extraction datasets that contain images and corresponding fully labeled building masks. Specifically, the newly constructed building labels can be randomly assigned according to the existing building masks, and the removed building labels can be generated within the background according to certain constraints.

Usually, building polygons are annotated manually based on ortho-images, with boundaries corresponding to building’s footprints. However, for large-scale building change monitoring, satellite images are a cost-effective option due to their quick and efficient acquisition. One of the challenges of using satellite images is the diverse misalignment between the building representation in images and their historical footprints, which is caused by the sensor perspective. The DCN-based feature transformation module of BCE-Net significantly relieves these inconsistencies. Nevertheless, it remains highly challenging to deal with the tremendous off-nadir problems of super-high-rise buildings. Additionally, the robustness of the BCE-Net in large-scale areas is open since there is no sufficient public data for support.

\section{Conclusions}
\label{s:conclusion}
We transformed a general building-change detection approach based on two-period images into a multi-task building segmentation framework, BCE-Net, which combines single up-to-date images with historical building masks to extract newly constructed buildings and removed buildings simultaneously. The advantages of BCE-Net are as follows. First, BCE-Net avoids the negative effects of pseudo-changes caused by variations in season and illumination between paired images. Second, BCE-Net incorporates a self-designed DCN-based feature transform module that further alleviates the off-nadir problem between buildings in images and historical masks. Third, as BCE-Net focuses on buildings, it sufficiently exploits latent consistency through instance-level contrastive learning for robust feature extraction and uses a random label-augmentation strategy to alleviate the unbalanced change labels.
 
Furthermore, we constructed a public dataset for building change extraction. This dataset consists of single temporal imagery and fully annotated change labels based on specific historical building polygons. We evaluated the performance BCE-Net and provide a new benchmark that considers the off-nadir problem of high-rise buildings in urban areas.

\section*{Acknowledgments}
\label{s:acknowledgments}
This work was supported in part by the National Natural Science Foundation of China (Projects No. 42230102, 42071355, and 41871291), the Sichuan Science and Technology Fund for Distinguished Young Scholars (22JCQN0110), and the Cultivation Program for the Excellent Doctoral Dissertation of Southwest Jiaotong University (2020YBPY09).

\bibliographystyle{model2-names}
\bibliography{bce_net}

\begin{thebibliography}{65}
\expandafter\ifx\csname natexlab\endcsname\relax\def\natexlab#1{#1}\fi
\providecommand{\url}[1]{\texttt{#1}}
\providecommand{\href}[2]{#2}
\providecommand{\path}[1]{#1}
\providecommand{\DOIprefix}{doi:}
\providecommand{\ArXivprefix}{arXiv:}
\providecommand{\URLprefix}{URL: }
\providecommand{\Pubmedprefix}{pmid:}
\providecommand{\doi}[1]{\href{http://dx.doi.org/#1}{\path{#1}}}
\providecommand{\Pubmed}[1]{\href{pmid:#1}{\path{#1}}}
\providecommand{\bibinfo}[2]{#2}
\ifx\xfnm\relax \def\xfnm[#1]{\unskip,\space#1}\fi
\bibitem[{{AccuCities}(2022)}]{accucities_3d_2022}
\bibinfo{author}{{AccuCities}}, \bibinfo{year}{2022}.
\newblock \bibinfo{title}{3d models of london}.
\newblock \URLprefix \url{www.accucities.com}.
\bibitem[{Afaq and Manocha(2021)}]{afaq2021analysis}
\bibinfo{author}{Afaq, Y.}, \bibinfo{author}{Manocha, A.},
  \bibinfo{year}{2021}.
\newblock \bibinfo{title}{Analysis on change detection techniques for remote
  sensing applications: A review}.
\newblock \bibinfo{journal}{Ecological Informatics} \bibinfo{volume}{63},
  \bibinfo{pages}{101310}.
\bibitem[{Bai et~al.(2021)Bai, Fu, Lu and Li}]{bai2021edge}
\bibinfo{author}{Bai, B.}, \bibinfo{author}{Fu, W.}, \bibinfo{author}{Lu, T.},
  \bibinfo{author}{Li, S.}, \bibinfo{year}{2021}.
\newblock \bibinfo{title}{Edge-guided recurrent convolutional neural network
  for multitemporal remote sensing image building change detection}.
\newblock \bibinfo{journal}{IEEE Transactions on Geoscience and Remote Sensing}
  \bibinfo{volume}{60}, \bibinfo{pages}{1--13}.
\bibitem[{Chen et~al.(2021a)Chen, Li and Shi}]{chen2021adversarial}
\bibinfo{author}{Chen, H.}, \bibinfo{author}{Li, W.}, \bibinfo{author}{Shi,
  Z.}, \bibinfo{year}{2021}a.
\newblock \bibinfo{title}{Adversarial instance augmentation for building change
  detection in remote sensing images}.
\newblock \bibinfo{journal}{IEEE Transactions on Geoscience and Remote Sensing}
  \bibinfo{volume}{60}, \bibinfo{pages}{1--16}.
\bibitem[{Chen et~al.(2021b)Chen, Qi and Shi}]{chen2021remote}
\bibinfo{author}{Chen, H.}, \bibinfo{author}{Qi, Z.}, \bibinfo{author}{Shi,
  Z.}, \bibinfo{year}{2021}b.
\newblock \bibinfo{title}{Remote sensing image change detection with
  transformers}.
\newblock \bibinfo{journal}{IEEE Transactions on Geoscience and Remote Sensing}
  \bibinfo{volume}{60}, \bibinfo{pages}{1--14}.
\bibitem[{Chen and Shi(2020)}]{chen2020spatial}
\bibinfo{author}{Chen, H.}, \bibinfo{author}{Shi, Z.}, \bibinfo{year}{2020}.
\newblock \bibinfo{title}{A spatial-temporal attention-based method and a new
  dataset for remote sensing image change detection}.
\newblock \bibinfo{journal}{Remote Sensing} \bibinfo{volume}{12},
  \bibinfo{pages}{1662}.
\bibitem[{Chen et~al.(2020a)Chen, Yuan, Peng, Chen, Huang, Zhu, Liu and
  Li}]{chen2020dasnet}
\bibinfo{author}{Chen, J.}, \bibinfo{author}{Yuan, Z.}, \bibinfo{author}{Peng,
  J.}, \bibinfo{author}{Chen, L.}, \bibinfo{author}{Huang, H.},
  \bibinfo{author}{Zhu, J.}, \bibinfo{author}{Liu, Y.}, \bibinfo{author}{Li,
  H.}, \bibinfo{year}{2020}a.
\newblock \bibinfo{title}{Dasnet: Dual attentive fully convolutional siamese
  networks for change detection in high-resolution satellite images}.
\newblock \bibinfo{journal}{IEEE Journal of Selected Topics in Applied Earth
  Observations and Remote Sensing} \bibinfo{volume}{14},
  \bibinfo{pages}{1194--1206}.
\bibitem[{Chen et~al.(2022a)Chen, Zhang, Hong, Chen, Yang and
  Li}]{chen2022fccdn}
\bibinfo{author}{Chen, P.}, \bibinfo{author}{Zhang, B.}, \bibinfo{author}{Hong,
  D.}, \bibinfo{author}{Chen, Z.}, \bibinfo{author}{Yang, X.},
  \bibinfo{author}{Li, B.}, \bibinfo{year}{2022}a.
\newblock \bibinfo{title}{Fccdn: Feature constraint network for vhr image
  change detection}.
\newblock \bibinfo{journal}{ISPRS Journal of Photogrammetry and Remote Sensing}
  \bibinfo{volume}{187}, \bibinfo{pages}{101--119}.
\bibitem[{Chen et~al.(2020b)Chen, Kornblith, Norouzi and
  Hinton}]{chen2020simple}
\bibinfo{author}{Chen, T.}, \bibinfo{author}{Kornblith, S.},
  \bibinfo{author}{Norouzi, M.}, \bibinfo{author}{Hinton, G.},
  \bibinfo{year}{2020}b.
\newblock \bibinfo{title}{A simple framework for contrastive learning of visual
  representations}, in: \bibinfo{booktitle}{International conference on machine
  learning}, \bibinfo{organization}{PMLR}. pp. \bibinfo{pages}{1597--1607}.
\bibitem[{Chen et~al.(2022b)Chen, Zhou, Wang, Xu, He, Jin and
  Jin}]{chen2022egde}
\bibinfo{author}{Chen, Z.}, \bibinfo{author}{Zhou, Y.}, \bibinfo{author}{Wang,
  B.}, \bibinfo{author}{Xu, X.}, \bibinfo{author}{He, N.},
  \bibinfo{author}{Jin, S.}, \bibinfo{author}{Jin, S.}, \bibinfo{year}{2022}b.
\newblock \bibinfo{title}{Egde-net: A building change detection method for
  high-resolution remote sensing imagery based on edge guidance and
  differential enhancement}.
\newblock \bibinfo{journal}{ISPRS Journal of Photogrammetry and Remote Sensing}
  \bibinfo{volume}{191}, \bibinfo{pages}{203--222}.
\bibitem[{Dai et~al.(2017)Dai, Qi, Xiong, Li, Zhang, Hu and
  Wei}]{dai2017deformable}
\bibinfo{author}{Dai, J.}, \bibinfo{author}{Qi, H.}, \bibinfo{author}{Xiong,
  Y.}, \bibinfo{author}{Li, Y.}, \bibinfo{author}{Zhang, G.},
  \bibinfo{author}{Hu, H.}, \bibinfo{author}{Wei, Y.}, \bibinfo{year}{2017}.
\newblock \bibinfo{title}{Deformable convolutional networks}, in:
  \bibinfo{booktitle}{Proceedings of the IEEE international conference on
  computer vision}, pp. \bibinfo{pages}{764--773}.
\bibitem[{Daudt et~al.(2018)Daudt, Le~Saux and Boulch}]{daudt2018fully}
\bibinfo{author}{Daudt, R.C.}, \bibinfo{author}{Le~Saux, B.},
  \bibinfo{author}{Boulch, A.}, \bibinfo{year}{2018}.
\newblock \bibinfo{title}{Fully convolutional siamese networks for change
  detection}, in: \bibinfo{booktitle}{2018 25th IEEE International Conference
  on Image Processing (ICIP)}, \bibinfo{organization}{IEEE}. pp.
  \bibinfo{pages}{4063--4067}.
\bibitem[{Ding et~al.(2021)Ding, Shao, Huang and Altan}]{ding2021dsa}
\bibinfo{author}{Ding, Q.}, \bibinfo{author}{Shao, Z.}, \bibinfo{author}{Huang,
  X.}, \bibinfo{author}{Altan, O.}, \bibinfo{year}{2021}.
\newblock \bibinfo{title}{Dsa-net: A novel deeply supervised attention-guided
  network for building change detection in high-resolution remote sensing
  images}.
\newblock \bibinfo{journal}{International Journal of Applied Earth Observation
  and Geoinformation} \bibinfo{volume}{105}, \bibinfo{pages}{102591}.
\bibitem[{Dong et~al.(2021)Dong, Zhao and Wang}]{dong2021multiscale}
\bibinfo{author}{Dong, J.}, \bibinfo{author}{Zhao, W.}, \bibinfo{author}{Wang,
  S.}, \bibinfo{year}{2021}.
\newblock \bibinfo{title}{Multiscale context aggregation network for building
  change detection using high resolution remote sensing images}.
\newblock \bibinfo{journal}{IEEE geoscience and remote sensing letters}
  \bibinfo{volume}{19}, \bibinfo{pages}{1--5}.
\bibitem[{Fan et~al.(2014)Fan, Zipf, Fu and Neis}]{fan2014quality}
\bibinfo{author}{Fan, H.}, \bibinfo{author}{Zipf, A.}, \bibinfo{author}{Fu,
  Q.}, \bibinfo{author}{Neis, P.}, \bibinfo{year}{2014}.
\newblock \bibinfo{title}{Quality assessment for building footprints data on
  openstreetmap}.
\newblock \bibinfo{journal}{International Journal of Geographical Information
  Science} \bibinfo{volume}{28}, \bibinfo{pages}{700--719}.
\bibitem[{Franklin et~al.(2011)Franklin, Kossmann, Kraska, Ramesh and
  Xin}]{franklin2011crowddb}
\bibinfo{author}{Franklin, M.J.}, \bibinfo{author}{Kossmann, D.},
  \bibinfo{author}{Kraska, T.}, \bibinfo{author}{Ramesh, S.},
  \bibinfo{author}{Xin, R.}, \bibinfo{year}{2011}.
\newblock \bibinfo{title}{Crowddb: answering queries with crowdsourcing}, in:
  \bibinfo{booktitle}{Proceedings of the 2011 ACM SIGMOD International
  Conference on Management of data}, pp. \bibinfo{pages}{61--72}.
\bibitem[{Gao et~al.(2022)Gao, Li, Sun, Wei, Chen and Wang}]{gao2022built}
\bibinfo{author}{Gao, S.}, \bibinfo{author}{Li, W.}, \bibinfo{author}{Sun, K.},
  \bibinfo{author}{Wei, J.}, \bibinfo{author}{Chen, Y.}, \bibinfo{author}{Wang,
  X.}, \bibinfo{year}{2022}.
\newblock \bibinfo{title}{Built-up area change detection using multi-task
  network with object-level refinement}.
\newblock \bibinfo{journal}{Remote Sensing} \bibinfo{volume}{14},
  \bibinfo{pages}{957}.
\bibitem[{Girard et~al.(2018)Girard, Charpiat and
  Tarabalka}]{girard2018aligning}
\bibinfo{author}{Girard, N.}, \bibinfo{author}{Charpiat, G.},
  \bibinfo{author}{Tarabalka, Y.}, \bibinfo{year}{2018}.
\newblock \bibinfo{title}{Aligning and updating cadaster maps with aerial
  images by multi-task, multi-resolution deep learning}, in:
  \bibinfo{booktitle}{Asian Conference on Computer Vision},
  \bibinfo{organization}{Springer}. pp. \bibinfo{pages}{675--690}.
\bibitem[{Goodfellow et~al.(2020)Goodfellow, Pouget-Abadie, Mirza, Xu,
  Warde-Farley, Ozair, Courville and Bengio}]{goodfellow2020generative}
\bibinfo{author}{Goodfellow, I.}, \bibinfo{author}{Pouget-Abadie, J.},
  \bibinfo{author}{Mirza, M.}, \bibinfo{author}{Xu, B.},
  \bibinfo{author}{Warde-Farley, D.}, \bibinfo{author}{Ozair, S.},
  \bibinfo{author}{Courville, A.}, \bibinfo{author}{Bengio, Y.},
  \bibinfo{year}{2020}.
\newblock \bibinfo{title}{Generative adversarial networks}.
\newblock \bibinfo{journal}{Communications of the ACM} \bibinfo{volume}{63},
  \bibinfo{pages}{139--144}.
\bibitem[{Guo et~al.(2018)Guo, Haque, Huang, Yeung and
  Fei-Fei}]{guo2018dynamic}
\bibinfo{author}{Guo, M.}, \bibinfo{author}{Haque, A.}, \bibinfo{author}{Huang,
  D.A.}, \bibinfo{author}{Yeung, S.}, \bibinfo{author}{Fei-Fei, L.},
  \bibinfo{year}{2018}.
\newblock \bibinfo{title}{Dynamic task prioritization for multitask learning},
  in: \bibinfo{booktitle}{Proceedings of the European conference on computer
  vision (ECCV)}, pp. \bibinfo{pages}{270--287}.
\bibitem[{Guo and Du(2017)}]{guo2017mining}
\bibinfo{author}{Guo, Z.}, \bibinfo{author}{Du, S.}, \bibinfo{year}{2017}.
\newblock \bibinfo{title}{Mining parameter information for building extraction
  and change detection with very high-resolution imagery and gis data}.
\newblock \bibinfo{journal}{GIScience \& Remote Sensing} \bibinfo{volume}{54},
  \bibinfo{pages}{38--63}.
\bibitem[{He et~al.(2022a)He, Sun, Diao, Yan, Yin and Fu}]{he2022transformer}
\bibinfo{author}{He, Q.}, \bibinfo{author}{Sun, X.}, \bibinfo{author}{Diao,
  W.}, \bibinfo{author}{Yan, Z.}, \bibinfo{author}{Yin, D.},
  \bibinfo{author}{Fu, K.}, \bibinfo{year}{2022}a.
\newblock \bibinfo{title}{Transformer-induced graph reasoning for multimodal
  semantic segmentation in remote sensing}.
\newblock \bibinfo{journal}{ISPRS Journal of Photogrammetry and Remote Sensing}
  \bibinfo{volume}{193}, \bibinfo{pages}{90--103}.
\bibitem[{He et~al.(2022b)He, Sun, Yan, Li and Fu}]{he2022multi}
\bibinfo{author}{He, Q.}, \bibinfo{author}{Sun, X.}, \bibinfo{author}{Yan, Z.},
  \bibinfo{author}{Li, B.}, \bibinfo{author}{Fu, K.}, \bibinfo{year}{2022}b.
\newblock \bibinfo{title}{Multi-object tracking in satellite videos with
  graph-based multitask modeling}.
\newblock \bibinfo{journal}{IEEE Transactions on Geoscience and Remote Sensing}
  \bibinfo{volume}{60}, \bibinfo{pages}{1--13}.
\bibitem[{Hu et~al.(2021)Hu, Lu and Ji}]{hu2021cascaded}
\bibinfo{author}{Hu, M.}, \bibinfo{author}{Lu, M.}, \bibinfo{author}{Ji, S.},
  \bibinfo{year}{2021}.
\newblock \bibinfo{title}{Cascaded deep neural networks for predicting biases
  between building polygons in vector maps and new remote sensing images}, in:
  \bibinfo{booktitle}{2021 IEEE International Geoscience and Remote Sensing
  Symposium IGARSS}, \bibinfo{organization}{IEEE}. pp.
  \bibinfo{pages}{4051--4054}.
\bibitem[{Huang et~al.(2021)Huang, Shen, Wang and Yang}]{huang2021multiple}
\bibinfo{author}{Huang, J.}, \bibinfo{author}{Shen, Q.}, \bibinfo{author}{Wang,
  M.}, \bibinfo{author}{Yang, M.}, \bibinfo{year}{2021}.
\newblock \bibinfo{title}{Multiple attention siamese network for
  high-resolution image change detection}.
\newblock \bibinfo{journal}{IEEE Transactions on Geoscience and Remote Sensing}
  \bibinfo{volume}{60}, \bibinfo{pages}{1--16}.
\bibitem[{Ji et~al.(2018)Ji, Wei and Lu}]{ji2018fully}
\bibinfo{author}{Ji, S.}, \bibinfo{author}{Wei, S.}, \bibinfo{author}{Lu, M.},
  \bibinfo{year}{2018}.
\newblock \bibinfo{title}{Fully convolutional networks for multisource building
  extraction from an open aerial and satellite imagery data set}.
\newblock \bibinfo{journal}{IEEE Transactions on Geoscience and Remote Sensing}
  \bibinfo{volume}{57}, \bibinfo{pages}{574--586}.
\bibitem[{Kang et~al.(2022)Kang, Wang, Zhu, Xia, Sun, Fernandez-Beltran and
  Plaza}]{kang2022disoptnet}
\bibinfo{author}{Kang, J.}, \bibinfo{author}{Wang, Z.}, \bibinfo{author}{Zhu,
  R.}, \bibinfo{author}{Xia, J.}, \bibinfo{author}{Sun, X.},
  \bibinfo{author}{Fernandez-Beltran, R.}, \bibinfo{author}{Plaza, A.},
  \bibinfo{year}{2022}.
\newblock \bibinfo{title}{Disoptnet: Distilling semantic knowledge from optical
  images for weather-independent building segmentation}.
\newblock \bibinfo{journal}{IEEE Transactions on Geoscience and Remote Sensing}
  \bibinfo{volume}{60}, \bibinfo{pages}{1--15}.
\bibitem[{Kendall et~al.(2018)Kendall, Gal and Cipolla}]{kendall2018multi}
\bibinfo{author}{Kendall, A.}, \bibinfo{author}{Gal, Y.},
  \bibinfo{author}{Cipolla, R.}, \bibinfo{year}{2018}.
\newblock \bibinfo{title}{Multi-task learning using uncertainty to weigh losses
  for scene geometry and semantics}, in: \bibinfo{booktitle}{Proceedings of the
  IEEE conference on computer vision and pattern recognition}, pp.
  \bibinfo{pages}{7482--7491}.
\bibitem[{Khosla et~al.(2020)Khosla, Teterwak, Wang, Sarna, Tian, Isola,
  Maschinot, Liu and Krishnan}]{khosla2020supervised}
\bibinfo{author}{Khosla, P.}, \bibinfo{author}{Teterwak, P.},
  \bibinfo{author}{Wang, C.}, \bibinfo{author}{Sarna, A.},
  \bibinfo{author}{Tian, Y.}, \bibinfo{author}{Isola, P.},
  \bibinfo{author}{Maschinot, A.}, \bibinfo{author}{Liu, C.},
  \bibinfo{author}{Krishnan, D.}, \bibinfo{year}{2020}.
\newblock \bibinfo{title}{Supervised contrastive learning}.
\newblock \bibinfo{journal}{Advances in Neural Information Processing Systems}
  \bibinfo{volume}{33}, \bibinfo{pages}{18661--18673}.
\bibitem[{{Lands Department}(2022)}]{lands_department_3d_2022}
\bibinfo{author}{{Lands Department}}, \bibinfo{year}{2022}.
\newblock \bibinfo{title}{3d spatial data}.
\newblock \URLprefix
  \url{https://data.gov.hk/sc-data/dataset/hk-landsd-openmap-development-hkms-digital-3d-bit00}.
\bibitem[{Li et~al.(2020)Li, Huang and Chang}]{li2020label}
\bibinfo{author}{Li, J.}, \bibinfo{author}{Huang, X.}, \bibinfo{author}{Chang,
  X.}, \bibinfo{year}{2020}.
\newblock \bibinfo{title}{A label-noise robust active learning sample
  collection method for multi-temporal urban land-cover classification and
  change analysis}.
\newblock \bibinfo{journal}{ISPRS Journal of Photogrammetry and Remote Sensing}
  \bibinfo{volume}{163}, \bibinfo{pages}{1--17}.
\bibitem[{Li et~al.(2022)Li, Mou, Hua, Shi and Zhu}]{li2022crossgeonet}
\bibinfo{author}{Li, Q.}, \bibinfo{author}{Mou, L.}, \bibinfo{author}{Hua, Y.},
  \bibinfo{author}{Shi, Y.}, \bibinfo{author}{Zhu, X.X.}, \bibinfo{year}{2022}.
\newblock \bibinfo{title}{Crossgeonet: A framework for building footprint
  generation of label-scarce geographical regions}.
\newblock \bibinfo{journal}{International Journal of Applied Earth Observation
  and Geoinformation} \bibinfo{volume}{111}, \bibinfo{pages}{102824}.
\bibitem[{Liu et~al.(2022)Liu, Xuan, Gan, Zhan, Liu and Du}]{liu2022end}
\bibinfo{author}{Liu, J.}, \bibinfo{author}{Xuan, W.}, \bibinfo{author}{Gan,
  Y.}, \bibinfo{author}{Zhan, Y.}, \bibinfo{author}{Liu, J.},
  \bibinfo{author}{Du, B.}, \bibinfo{year}{2022}.
\newblock \bibinfo{title}{An end-to-end supervised domain adaptation framework
  for cross-domain change detection}.
\newblock \bibinfo{journal}{Pattern Recognition} \bibinfo{volume}{132},
  \bibinfo{pages}{108960}.
\bibitem[{Liu et~al.(2021a)Liu, Shi, Marinoni, He, Liu and
  Zhang}]{liu2021super}
\bibinfo{author}{Liu, M.}, \bibinfo{author}{Shi, Q.},
  \bibinfo{author}{Marinoni, A.}, \bibinfo{author}{He, D.},
  \bibinfo{author}{Liu, X.}, \bibinfo{author}{Zhang, L.},
  \bibinfo{year}{2021}a.
\newblock \bibinfo{title}{Super-resolution-based change detection network with
  stacked attention module for images with different resolutions}.
\newblock \bibinfo{journal}{IEEE Transactions on Geoscience and Remote Sensing}
  \bibinfo{volume}{60}, \bibinfo{pages}{1--18}.
\bibitem[{Liu et~al.(2020a)Liu, Jiang, Zhang and Zhang}]{liu2020deep}
\bibinfo{author}{Liu, R.}, \bibinfo{author}{Jiang, D.}, \bibinfo{author}{Zhang,
  L.}, \bibinfo{author}{Zhang, Z.}, \bibinfo{year}{2020}a.
\newblock \bibinfo{title}{Deep depthwise separable convolutional network for
  change detection in optical aerial images}.
\newblock \bibinfo{journal}{IEEE Journal of Selected Topics in Applied Earth
  Observations and Remote Sensing} \bibinfo{volume}{13},
  \bibinfo{pages}{1109--1118}.
\bibitem[{Liu et~al.(2021b)Liu, Gong, Lu, Zhang, Zheng, Jiang and
  Zhang}]{liu2021building}
\bibinfo{author}{Liu, T.}, \bibinfo{author}{Gong, M.}, \bibinfo{author}{Lu,
  D.}, \bibinfo{author}{Zhang, Q.}, \bibinfo{author}{Zheng, H.},
  \bibinfo{author}{Jiang, F.}, \bibinfo{author}{Zhang, M.},
  \bibinfo{year}{2021}b.
\newblock \bibinfo{title}{Building change detection for vhr remote sensing
  images via local--global pyramid network and cross-task transfer learning
  strategy}.
\newblock \bibinfo{journal}{IEEE Transactions on Geoscience and Remote Sensing}
  \bibinfo{volume}{60}, \bibinfo{pages}{1--17}.
\bibitem[{Liu et~al.(2020b)Liu, Pang, Zhan, Zhang and Yang}]{liu2020building}
\bibinfo{author}{Liu, Y.}, \bibinfo{author}{Pang, C.}, \bibinfo{author}{Zhan,
  Z.}, \bibinfo{author}{Zhang, X.}, \bibinfo{author}{Yang, X.},
  \bibinfo{year}{2020}b.
\newblock \bibinfo{title}{Building change detection for remote sensing images
  using a dual-task constrained deep siamese convolutional network model}.
\newblock \bibinfo{journal}{IEEE Geoscience and Remote Sensing Letters}
  \bibinfo{volume}{18}, \bibinfo{pages}{811--815}.
\bibitem[{Long et~al.(2015)Long, Shelhamer and Darrell}]{long2015fully}
\bibinfo{author}{Long, J.}, \bibinfo{author}{Shelhamer, E.},
  \bibinfo{author}{Darrell, T.}, \bibinfo{year}{2015}.
\newblock \bibinfo{title}{Fully convolutional networks for semantic
  segmentation}, in: \bibinfo{booktitle}{Proceedings of the IEEE conference on
  computer vision and pattern recognition}, pp. \bibinfo{pages}{3431--3440}.
\bibitem[{Lv et~al.(2020)Lv, Liu and Benediktsson}]{lv2020object}
\bibinfo{author}{Lv, Z.}, \bibinfo{author}{Liu, T.},
  \bibinfo{author}{Benediktsson, J.A.}, \bibinfo{year}{2020}.
\newblock \bibinfo{title}{Object-oriented key point vector distance for binary
  land cover change detection using vhr remote sensing images}.
\newblock \bibinfo{journal}{IEEE Transactions on Geoscience and Remote Sensing}
  \bibinfo{volume}{58}, \bibinfo{pages}{6524--6533}.
\bibitem[{Lv et~al.(2021)Lv, Liu, Benediktsson and Falco}]{lv2021land}
\bibinfo{author}{Lv, Z.}, \bibinfo{author}{Liu, T.},
  \bibinfo{author}{Benediktsson, J.A.}, \bibinfo{author}{Falco, N.},
  \bibinfo{year}{2021}.
\newblock \bibinfo{title}{Land cover change detection techniques:
  Very-high-resolution optical images: A review}.
\newblock \bibinfo{journal}{IEEE Geoscience and Remote Sensing Magazine}
  \bibinfo{volume}{10}, \bibinfo{pages}{44--63}.
\bibitem[{Pan and Yang(2009)}]{pan2009survey}
\bibinfo{author}{Pan, S.J.}, \bibinfo{author}{Yang, Q.}, \bibinfo{year}{2009}.
\newblock \bibinfo{title}{A survey on transfer learning}.
\newblock \bibinfo{journal}{IEEE Transactions on knowledge and data
  engineering} \bibinfo{volume}{22}, \bibinfo{pages}{1345--1359}.
\bibitem[{Peng et~al.(2020a)Peng, Bruzzone, Zhang, Guan, Ding and
  Huang}]{peng2020semicdnet}
\bibinfo{author}{Peng, D.}, \bibinfo{author}{Bruzzone, L.},
  \bibinfo{author}{Zhang, Y.}, \bibinfo{author}{Guan, H.},
  \bibinfo{author}{Ding, H.}, \bibinfo{author}{Huang, X.},
  \bibinfo{year}{2020}a.
\newblock \bibinfo{title}{Semicdnet: A semisupervised convolutional neural
  network for change detection in high resolution remote-sensing images}.
\newblock \bibinfo{journal}{IEEE Transactions on Geoscience and Remote Sensing}
  \bibinfo{volume}{59}, \bibinfo{pages}{5891--5906}.
\bibitem[{Peng et~al.(2019)Peng, Zhang and Guan}]{peng2019end}
\bibinfo{author}{Peng, D.}, \bibinfo{author}{Zhang, Y.}, \bibinfo{author}{Guan,
  H.}, \bibinfo{year}{2019}.
\newblock \bibinfo{title}{End-to-end change detection for high resolution
  satellite images using improved unet++}.
\newblock \bibinfo{journal}{Remote Sensing} \bibinfo{volume}{11},
  \bibinfo{pages}{1382}.
\bibitem[{Peng et~al.(2020b)Peng, Zhong, Li and Li}]{peng2020optical}
\bibinfo{author}{Peng, X.}, \bibinfo{author}{Zhong, R.}, \bibinfo{author}{Li,
  Z.}, \bibinfo{author}{Li, Q.}, \bibinfo{year}{2020}b.
\newblock \bibinfo{title}{Optical remote sensing image change detection based
  on attention mechanism and image difference}.
\newblock \bibinfo{journal}{IEEE Transactions on Geoscience and Remote Sensing}
  \bibinfo{volume}{59}, \bibinfo{pages}{7296--7307}.
\bibitem[{Ronneberger et~al.(2015)Ronneberger, Fischer and
  Brox}]{ronneberger2015u}
\bibinfo{author}{Ronneberger, O.}, \bibinfo{author}{Fischer, P.},
  \bibinfo{author}{Brox, T.}, \bibinfo{year}{2015}.
\newblock \bibinfo{title}{U-net: Convolutional networks for biomedical image
  segmentation}, in: \bibinfo{booktitle}{International Conference on Medical
  image computing and computer-assisted intervention},
  \bibinfo{organization}{Springer}. pp. \bibinfo{pages}{234--241}.
\bibitem[{Sefrin et~al.(2020)Sefrin, Riese and Keller}]{sefrin2020deep}
\bibinfo{author}{Sefrin, O.}, \bibinfo{author}{Riese, F.M.},
  \bibinfo{author}{Keller, S.}, \bibinfo{year}{2020}.
\newblock \bibinfo{title}{Deep learning for land cover change detection}.
\newblock \bibinfo{journal}{Remote Sensing} \bibinfo{volume}{13},
  \bibinfo{pages}{78}.
\bibitem[{Shen et~al.(2022)Shen, Huang, Wang, Tao, Yang and
  Zhang}]{shen2022semantic}
\bibinfo{author}{Shen, Q.}, \bibinfo{author}{Huang, J.}, \bibinfo{author}{Wang,
  M.}, \bibinfo{author}{Tao, S.}, \bibinfo{author}{Yang, R.},
  \bibinfo{author}{Zhang, X.}, \bibinfo{year}{2022}.
\newblock \bibinfo{title}{Semantic feature-constrained multitask siamese
  network for building change detection in high-spatial-resolution remote
  sensing imagery}.
\newblock \bibinfo{journal}{ISPRS Journal of Photogrammetry and Remote Sensing}
  \bibinfo{volume}{189}, \bibinfo{pages}{78--94}.
\bibitem[{Shi et~al.(2020)Shi, Zhang, Zhang, Chen and Zhan}]{shi2020change}
\bibinfo{author}{Shi, W.}, \bibinfo{author}{Zhang, M.}, \bibinfo{author}{Zhang,
  R.}, \bibinfo{author}{Chen, S.}, \bibinfo{author}{Zhan, Z.},
  \bibinfo{year}{2020}.
\newblock \bibinfo{title}{Change detection based on artificial intelligence:
  State-of-the-art and challenges}.
\newblock \bibinfo{journal}{Remote Sensing} \bibinfo{volume}{12},
  \bibinfo{pages}{1688}.
\bibitem[{Sun et~al.(2022)Sun, Wang, Yan, Xu, Wang, Diao, Chen, Li, Feng, Xu
  et~al.}]{sun2022fair1m}
\bibinfo{author}{Sun, X.}, \bibinfo{author}{Wang, P.}, \bibinfo{author}{Yan,
  Z.}, \bibinfo{author}{Xu, F.}, \bibinfo{author}{Wang, R.},
  \bibinfo{author}{Diao, W.}, \bibinfo{author}{Chen, J.}, \bibinfo{author}{Li,
  J.}, \bibinfo{author}{Feng, Y.}, \bibinfo{author}{Xu, T.}, et~al.,
  \bibinfo{year}{2022}.
\newblock \bibinfo{title}{Fair1m: A benchmark dataset for fine-grained object
  recognition in high-resolution remote sensing imagery}.
\newblock \bibinfo{journal}{ISPRS Journal of Photogrammetry and Remote Sensing}
  \bibinfo{volume}{184}, \bibinfo{pages}{116--130}.
\bibitem[{Sun et~al.(2020)Sun, Zhang, Huang, Wang and Xin}]{sun2020fine}
\bibinfo{author}{Sun, Y.}, \bibinfo{author}{Zhang, X.}, \bibinfo{author}{Huang,
  J.}, \bibinfo{author}{Wang, H.}, \bibinfo{author}{Xin, Q.},
  \bibinfo{year}{2020}.
\newblock \bibinfo{title}{Fine-grained building change detection from very
  high-spatial-resolution remote sensing images based on deep multitask
  learning}.
\newblock \bibinfo{journal}{IEEE Geoscience and Remote Sensing Letters}
  \bibinfo{volume}{19}, \bibinfo{pages}{1--5}.
\bibitem[{Vieira et~al.(2012)Vieira, Formaggio, Renn{\'o}, Atzberger, Aguiar
  and Mello}]{vieira2012object}
\bibinfo{author}{Vieira, M.A.}, \bibinfo{author}{Formaggio, A.R.},
  \bibinfo{author}{Renn{\'o}, C.D.}, \bibinfo{author}{Atzberger, C.},
  \bibinfo{author}{Aguiar, D.A.}, \bibinfo{author}{Mello, M.P.},
  \bibinfo{year}{2012}.
\newblock \bibinfo{title}{Object based image analysis and data mining applied
  to a remotely sensed landsat time-series to map sugarcane over large areas}.
\newblock \bibinfo{journal}{Remote Sensing of Environment}
  \bibinfo{volume}{123}, \bibinfo{pages}{553--562}.
\bibitem[{Wang et~al.(2022)Wang, Meng, Li, Yang, Yu and Xia}]{wang2022learning}
\bibinfo{author}{Wang, J.}, \bibinfo{author}{Meng, L.}, \bibinfo{author}{Li,
  W.}, \bibinfo{author}{Yang, W.}, \bibinfo{author}{Yu, L.},
  \bibinfo{author}{Xia, G.S.}, \bibinfo{year}{2022}.
\newblock \bibinfo{title}{Learning to extract building footprints from
  off-nadir aerial images}.
\newblock \bibinfo{journal}{arXiv preprint arXiv:2204.13637} .
\bibitem[{Yang et~al.(2021)Yang, Cao, Wan, Zhang and Tan}]{yang2021spatio}
\bibinfo{author}{Yang, Z.}, \bibinfo{author}{Cao, Z.}, \bibinfo{author}{Wan,
  X.}, \bibinfo{author}{Zhang, F.}, \bibinfo{author}{Tan, G.},
  \bibinfo{year}{2021}.
\newblock \bibinfo{title}{Spatio-temporal features processing network for
  change detection in remote sensing images}, in: \bibinfo{booktitle}{2021 IEEE
  International Geoscience and Remote Sensing Symposium IGARSS},
  \bibinfo{organization}{IEEE}. pp. \bibinfo{pages}{3344--3347}.
\bibitem[{Zhan et~al.(2017)Zhan, Fu, Yan, Sun, Wang and Qiu}]{zhan2017change}
\bibinfo{author}{Zhan, Y.}, \bibinfo{author}{Fu, K.}, \bibinfo{author}{Yan,
  M.}, \bibinfo{author}{Sun, X.}, \bibinfo{author}{Wang, H.},
  \bibinfo{author}{Qiu, X.}, \bibinfo{year}{2017}.
\newblock \bibinfo{title}{Change detection based on deep siamese convolutional
  network for optical aerial images}.
\newblock \bibinfo{journal}{IEEE Geoscience and Remote Sensing Letters}
  \bibinfo{volume}{14}, \bibinfo{pages}{1845--1849}.
\bibitem[{Zhang et~al.(2022)Zhang, Wang, Cheng and Li}]{zhang2022swinsunet}
\bibinfo{author}{Zhang, C.}, \bibinfo{author}{Wang, L.},
  \bibinfo{author}{Cheng, S.}, \bibinfo{author}{Li, Y.}, \bibinfo{year}{2022}.
\newblock \bibinfo{title}{Swinsunet: Pure transformer network for remote
  sensing image change detection}.
\newblock \bibinfo{journal}{IEEE Transactions on Geoscience and Remote Sensing}
  \bibinfo{volume}{60}, \bibinfo{pages}{1--13}.
\bibitem[{Zhang et~al.(2020)Zhang, Yue, Tapete, Jiang, Shangguan, Huang and
  Liu}]{zhang2020deeply}
\bibinfo{author}{Zhang, C.}, \bibinfo{author}{Yue, P.},
  \bibinfo{author}{Tapete, D.}, \bibinfo{author}{Jiang, L.},
  \bibinfo{author}{Shangguan, B.}, \bibinfo{author}{Huang, L.},
  \bibinfo{author}{Liu, G.}, \bibinfo{year}{2020}.
\newblock \bibinfo{title}{A deeply supervised image fusion network for change
  detection in high resolution bi-temporal remote sensing images}.
\newblock \bibinfo{journal}{ISPRS Journal of Photogrammetry and Remote Sensing}
  \bibinfo{volume}{166}, \bibinfo{pages}{183--200}.
\bibitem[{Zhang et~al.(2021a)Zhang, Hu, Zhang, Shu and Zhou}]{zhang2021object}
\bibinfo{author}{Zhang, L.}, \bibinfo{author}{Hu, X.}, \bibinfo{author}{Zhang,
  M.}, \bibinfo{author}{Shu, Z.}, \bibinfo{author}{Zhou, H.},
  \bibinfo{year}{2021}a.
\newblock \bibinfo{title}{Object-level change detection with a dual correlation
  attention-guided detector}.
\newblock \bibinfo{journal}{ISPRS Journal of Photogrammetry and Remote Sensing}
  \bibinfo{volume}{177}, \bibinfo{pages}{147--160}.
\bibitem[{Zhang et~al.(2017)Zhang, Xiao, Feng and Yuan}]{zhang2017separate}
\bibinfo{author}{Zhang, X.}, \bibinfo{author}{Xiao, P.}, \bibinfo{author}{Feng,
  X.}, \bibinfo{author}{Yuan, M.}, \bibinfo{year}{2017}.
\newblock \bibinfo{title}{Separate segmentation of multi-temporal
  high-resolution remote sensing images for object-based change detection in
  urban area}.
\newblock \bibinfo{journal}{Remote Sensing of Environment}
  \bibinfo{volume}{201}, \bibinfo{pages}{243--255}.
\bibitem[{Zhang et~al.(2021b)Zhang, Deng, He, Guo, Sun and
  Chen}]{zhang2021foda}
\bibinfo{author}{Zhang, Y.}, \bibinfo{author}{Deng, M.}, \bibinfo{author}{He,
  F.}, \bibinfo{author}{Guo, Y.}, \bibinfo{author}{Sun, G.},
  \bibinfo{author}{Chen, J.}, \bibinfo{year}{2021}b.
\newblock \bibinfo{title}{Foda: Building change detection in high-resolution
  remote sensing images based on feature--output space dual-alignment}.
\newblock \bibinfo{journal}{IEEE Journal of Selected Topics in Applied Earth
  Observations and Remote Sensing} \bibinfo{volume}{14},
  \bibinfo{pages}{8125--8134}.
\bibitem[{Zheng et~al.(2022)Zheng, Gong, Liu, Jiang, Zhan, Lu and
  Zhang}]{zheng2022hfa}
\bibinfo{author}{Zheng, H.}, \bibinfo{author}{Gong, M.}, \bibinfo{author}{Liu,
  T.}, \bibinfo{author}{Jiang, F.}, \bibinfo{author}{Zhan, T.},
  \bibinfo{author}{Lu, D.}, \bibinfo{author}{Zhang, M.}, \bibinfo{year}{2022}.
\newblock \bibinfo{title}{Hfa-net: High frequency attention siamese network for
  building change detection in vhr remote sensing images}.
\newblock \bibinfo{journal}{Pattern Recognition} \bibinfo{volume}{129},
  \bibinfo{pages}{108717}.
\bibitem[{Zheng et~al.(2021a)Zheng, Ma, Zhang and Zhong}]{zheng2021deep}
\bibinfo{author}{Zheng, Z.}, \bibinfo{author}{Ma, A.}, \bibinfo{author}{Zhang,
  L.}, \bibinfo{author}{Zhong, Y.}, \bibinfo{year}{2021}a.
\newblock \bibinfo{title}{Deep multisensor learning for missing-modality
  all-weather mapping}.
\newblock \bibinfo{journal}{ISPRS Journal of Photogrammetry and Remote Sensing}
  \bibinfo{volume}{174}, \bibinfo{pages}{254--264}.
\bibitem[{Zheng et~al.(2021b)Zheng, Wan, Zhang, Xiang, Peng and
  Zhang}]{zheng2021clnet}
\bibinfo{author}{Zheng, Z.}, \bibinfo{author}{Wan, Y.}, \bibinfo{author}{Zhang,
  Y.}, \bibinfo{author}{Xiang, S.}, \bibinfo{author}{Peng, D.},
  \bibinfo{author}{Zhang, B.}, \bibinfo{year}{2021}b.
\newblock \bibinfo{title}{Clnet: Cross-layer convolutional neural network for
  change detection in optical remote sensing imagery}.
\newblock \bibinfo{journal}{ISPRS Journal of Photogrammetry and Remote Sensing}
  \bibinfo{volume}{175}, \bibinfo{pages}{247--267}.
\bibitem[{Zhong et~al.(2015)Zhong, Xu, Yang and Hu}]{zhong2015building}
\bibinfo{author}{Zhong, C.}, \bibinfo{author}{Xu, Q.}, \bibinfo{author}{Yang,
  F.}, \bibinfo{author}{Hu, L.}, \bibinfo{year}{2015}.
\newblock \bibinfo{title}{Building change detection for high-resolution
  remotely sensed images based on a semantic dependency}, in:
  \bibinfo{booktitle}{2015 IEEE international geoscience and remote sensing
  symposium (IGARSS)}, \bibinfo{organization}{IEEE}. pp.
  \bibinfo{pages}{3345--3348}.
\bibitem[{Zhu et~al.(2021)Zhu, Huang, Hu, Li, Chen and Zhong}]{zhu2021depth}
\bibinfo{author}{Zhu, Q.}, \bibinfo{author}{Huang, S.}, \bibinfo{author}{Hu,
  H.}, \bibinfo{author}{Li, H.}, \bibinfo{author}{Chen, M.},
  \bibinfo{author}{Zhong, R.}, \bibinfo{year}{2021}.
\newblock \bibinfo{title}{Depth-enhanced feature pyramid network for
  occlusion-aware verification of buildings from oblique images}.
\newblock \bibinfo{journal}{ISPRS Journal of Photogrammetry and Remote Sensing}
  \bibinfo{volume}{174}, \bibinfo{pages}{105--116}.
\bibitem[{Zhu et~al.(2008)Zhu, Zhang and Pang}]{zhu2008change}
\bibinfo{author}{Zhu, X.}, \bibinfo{author}{Zhang, H.}, \bibinfo{author}{Pang,
  G.}, \bibinfo{year}{2008}.
\newblock \bibinfo{title}{A change detection method with high resolution images
  based on polygon automatic validating}.
\newblock \bibinfo{journal}{Int Arch Photogramm Remote Sens Spat Inf Sci
  Beijing China} \bibinfo{volume}{37}, \bibinfo{pages}{841--845}.

\end{thebibliography}

\end{document}